\documentclass{article}

\usepackage{microtype}
\usepackage{graphicx}
\usepackage{subfigure}
\usepackage{booktabs}

\usepackage{hyperref}

\usepackage[accepted]{mlsys2020}

\mlsystitlerunning{VS-Quant: Per-vector Scaled Quantization for Accurate Low-Precision Neural Network Inference}

\usepackage{enumitem}
\usepackage{mathtools}
\usepackage{multirow}
\usepackage[table]{xcolor}
\usepackage{array}
\usepackage{amsfonts}

\allowdisplaybreaks

\newcommand{\TT}[1]{\texttt{#1}}
\newcommand{\BF}[1]{\textbf{#1}}
\newcommand{\IT}[1]{\textit{#1}}

\begin{document}

\twocolumn[
\mlsystitle{VS-Quant: Per-vector Scaled Quantization for \\Accurate Low-Precision Neural Network Inference}

\begin{mlsysauthorlist}
\mlsysauthor{Steve Dai\hspace{6pt}}{}
\mlsysauthor{Rangharajan Venkatesan\hspace{6pt}}{}
\mlsysauthor{Haoxing Ren\hspace{6pt}}{}
\mlsysauthor{Brian Zimmer\hspace{6pt}}{}
\mlsysauthor{William J. Dally\hspace{6pt}}{}
\mlsysauthor{Brucek Khailany}{}
\\\vspace{5pt}NVIDIA
\end{mlsysauthorlist}

\mlsyskeywords{Machine Learning, MLSys}

\vskip 0.38in

\begin{abstract}

Quantization enables efficient acceleration of deep neural networks by reducing model memory footprint and exploiting low-cost integer math hardware units.
Quantization maps floating-point weights and activations in a trained model to low-bitwidth integer values using scale factors. Excessive quantization, reducing precision too aggressively, results in  accuracy degradation.  When scale factors are shared at a coarse granularity across many dimensions of each tensor, effective precision of individual elements within the tensor are limited.
To reduce quantization-related accuracy loss, we propose using a separate scale factor for each small vector of ($\approx$16-64) elements within a single dimension of a tensor.
To achieve an efficient hardware implementation, the per-vector scale factors can be implemented with low-bitwidth integers when calibrated using a two-level quantization scheme.  
We find that per-vector scaling consistently achieves better inference accuracy at low precision compared to conventional scaling techniques for popular neural networks without requiring retraining.
We also modify a deep learning accelerator hardware design to study the area and energy overheads of per-vector scaling support.
Our evaluation demonstrates that per-vector scaled quantization with 4-bit weights and activations achieves 37\% area saving and 24\% energy saving while maintaining over 75\% accuracy for ResNet50 on ImageNet.
4-bit weights and 8-bit activations achieve near-full-precision accuracy for both BERT-base and BERT-large on SQuAD while reducing area by 26\% compared to an 8-bit baseline.
\end{abstract}

]

\section{Introduction}

Deep neural networks (DNNs) continue to achieve groundbreaking accuracy on a range of tasks, including image classification, object detection, machine translation, and natural language processing (NLP)~\cite{lecun2015deep}.  In parallel, hardware designers have been racing to achieve the best performance per watt for running DNN inference on devices ranging from the edge to the datacenter~\cite{sze2020efficient}.
While most DNN models are trained in single-precision floating-point, they can be deployed for inference in lower-precision formats such as half-precision floating-point, fixed-point, and low-bitwidth integer depending on the target device and application specifications.
Quantizing DNN models to lower precisions allows us to accelerate compute-bound operations such as convolution on high-throughput low-cost math units, conserve memory bandwidth for memory-bound computations, and reduce storage requirements in on-chip buffers and caches.
For example, NVIDIA's Ampere Graphics Processing Unit (GPU) architecture supports INT8 and INT4 data types for these purposes~\cite{ampere3}.

One way to quantize a DNN model is through quantization-aware training (QAT).  QAT either trains the model from scratch or fine-tunes the trained full-precision model, with quantization operations included in the model.
Alternatively, post-training quantization (PTQ) directly quantizes the values of the full-precision model before and during inference without any retraining~\cite{wu2020integer}.
Often, PTQ is more desirable because it does not require access to the complete set of possibly confidential training data, eliminates lengthy fine-tuning, requires little hyperparameter tuning, and provides a turnkey solution for quantizing any DNN.
However, PTQ usually results in more accuracy loss than QAT because of the lack of training with quantizers in the loop.
With both QAT and PTQ, accuracy loss from quantization varies by precision, model, and quantization algorithm.

\begin{figure}[t]
\centering
\includegraphics[trim=0pt 380pt 380pt 0pt, width=\columnwidth]{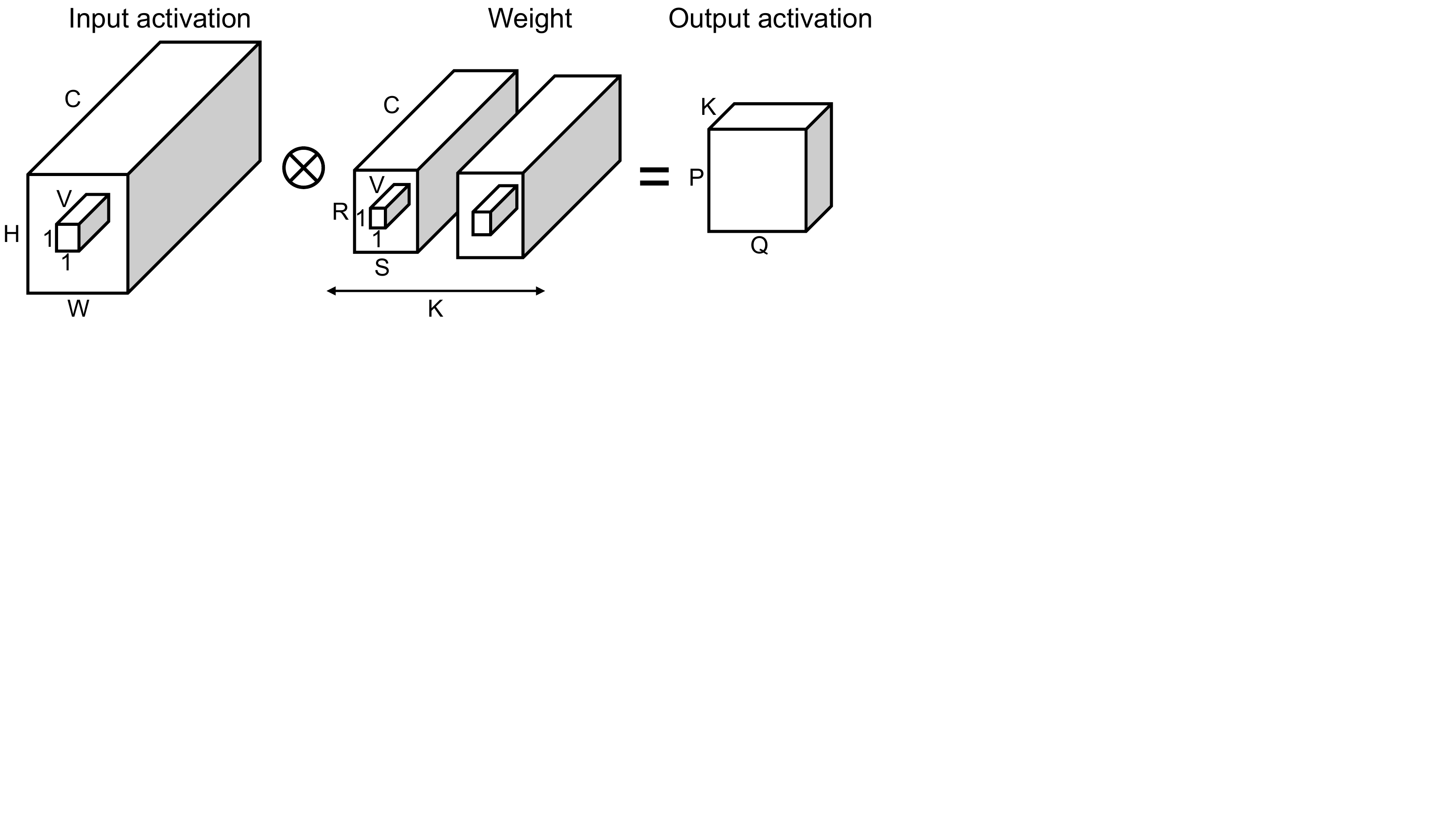}
\caption{\BF{Convolution} --- Comparison between per-layer/per-output-channel scaling and per-vector scaling.}\vspace{5pt}
\label{fig:convolution}
\end{figure}

Quantization scales high-precision values of a particular range to lower-precision values of a different range.
A high-precision number ($x$) is mapped to a lower-precision number ($x_q$) with $x_q=Q(x/s,N)$ where $s$ is the scale factor and $Q(a,b)$ is the function that quantizes $a$ to a $b$-bit integer.
Scale factors play an important role in determining the quantization error, which affects the ultimate accuracy of the quantized model.
To avoid overloading the quantized model with too many scale factors and nullifying the compute and memory benefits of quantization, scale factors must be shared among multiple tensor elements.
Typically, scale factors are shared at a coarse granularity by elements of an entire tensor or a large sub-tensor.
For example, a typical quantized convolution layer as shown in Figure~\ref{fig:convolution} employs a single scale factor for the entire input activation tensor ($C\times H\times W$) and each kernel ($C\times R\times S$) of the weight tensor.
While coarse-grained scaling amortizes the cost of scaling across many elements, it likely requires mapping a broader range of values to the specified low-precision representation.
The resulting increase in quantization error introduces significant accuracy loss for low-precision representations.
The problem is exacerbated for DNNs whose input and/or weight values span a wide dynamic range.

We propose fine-grained per-vector scaled quantization ({\em VS-Quant}) to mitigate quantization-related accuracy loss.
In contrast to coarse-grained per-layer/per-output-channel scaling, {\em VS-Quant} employs a scale factor for each vector of elements ($V\times 1\times 1$) in the activation and/or weight tensor as shown in Figure~\ref{fig:convolution}.
The range that must be represented within each vector is smaller than the range that must be represented
across the entire layer, so many vectors can be represented at much higher precision and quantization error is only encountered in a small subset of vectors that contain wide ranges of values.
Moreover, the unit of a vector matches the unit of vector multiply-and-accumulate (MAC) hardware ubiquitous in DNN accelerators~\cite{sijstermans2018nvdla,venkatesan2019magnet,ampere3}.
This hardware-software synergy leads to an elegant extension of current accelerator architectures for implementing per-vector scaling with low overhead.
The major contributions of our work are as follows:
\begin{itemize}[leftmargin=*]
\item We propose {\em VS-Quant}, a novel per-vector scaled quantization technique to mitigate accuracy loss typical in existing quantized DNN models.
\item We propose a two-level scaling scheme and algorithm that combine a set of fine-grained scale factors with each coarse-grained scale factor to enable efficient {\em VS-Quant} hardware implementations.
\item We evaluate {\em VS-Quant} on popular DNN models and demonstrate significantly higher PTQ accuracy than conventionally scaled quantization on computer vision and NLP tasks.  
\item We extend the vector MAC unit of a DNN accelerator to support {\em VS-Quant} in hardware and analyze the area and power impact.
\item We explore tradeoffs between accuracy and hardware efficiency across a range of hardware implementations and DNN models to identify Pareto-optimal designs for low-precision inference with PTQ.
\end{itemize}

The remainder of the paper is organized as follows: Section~\ref{sec-related} reviews related work;
Section~\ref{sec-quant} describes the fundamentals for quantization; 
Section~\ref{sec-pvsq} presents and evaluates our per-vector scaling technique and associated two-level scaling scheme; 
Section~\ref{sec-hw} describes the hardware implementation; 
Section~\ref{sec-dse} explores the accuracy and hardware efficiency tradeoff;
Section~\ref{sec-discussions} discusses quantization-aware retraining in the context of per-vector scaling, followed by conclusions in Section~\ref{sec-conclusions}.

\section{Related Work}
\label{sec-related}

Krishnamoorthi evaluates per-channel scaled quantization at various precisions for a set of convolutional neural networks (CNNs) ~\cite{krishnamoorthi2018quantizing}. 
The paper finds that although PTQ can achieve good accuracy at 8 bits for these networks, QAT is required for getting good accuracy at lower precisions or for matching floating-point accuracy at 8-bits.
McKinstry et al. shows that CNNs require only a small number of epochs of finetuning after carefully setting the learning rate schedule and fixing the quantization range~\cite{mckinstry2018discovering}.
Instead of fixing the quantization range before QAT, PACT proposes to learn the range of weights and activations as part of training~\cite{choi2018pact}.
Both papers achieve full-precision accuracy with only 4-bit precision.
Other research has explored very low precision ternary~\cite{zhu2016trained} and binary~\cite{courbariaux2015binaryconnect,hubara2016binarized} weights and activations.
These models required significant retraining to recover accuracy loss and do not reach full-precision accuracy for more difficult tasks. 
In addition to CNNs, recent work has proposed quantized transformer models for NLP~\cite{zafrir2019q8bert,shen2020q} and for machine translation~\cite{bhandare2019efficient,prato2019fully,wu2016google}.
Wu et al. establishes a single 8-bit quantization workflow for maintaining less than 1\% accuracy drop for many different types of networks~\cite{wu2020integer}.

Prior work has proposed schemes for uniform quantization~\cite{courbariaux2014training,zhou2016dorefa} and non-uniform quantization~\cite{han2015deep,zhu2016trained}.
Uniform quantization uses integer or fixed-point format which can be accelerated with specialized math pipelines and is the focus of this paper.
Non-uniform quantization leverages codebook look-ups to enable model compression and memory bandwidth reduction.
To reduce quantization error, vector quantization~\cite{gray1984vector} based techniques take advantage of redundancy within a subspace of a weight or activation tensor.
In particular, product quantization splits each subspace into vectors and optimizes a codebook for the vectors in each subspace~\cite{wu2016quantized,gong2014compressing}.
Stock et al. extends this technique to preserve the reconstruction quality of actual network outputs instead of just weights~\cite{stock2020and}.
On a separate note, knowledge distillation can also improve the accuracy of quantized model~\cite{mishra2017apprentice}.
By training the quantized model to mimic a high-precision model in a student-teacher setting, the paper obtains higher accuracy for a ternary ResNet-variant architecture.

Since the full set of training data may not be available at inference time, there is increasing interest in PTQ techniques that directly quantize full-precision values before and during inference~\cite{krishnamoorthi2018quantizing,lee2018quantization,nagel2019data}.
More recently, Zhao et al. proposes the outlier channel splitting technique to exactly represent outliers~\cite{zhao2019improving}.
By duplicating channels that contain outliers and halving the values of those channels, this technique effectively shrinks the quantization range without modifying the network.
Also focusing on the distribution of tensor values, Fang et al. proposes a piecewise linear quantization scheme that optimally splits the quantization range into non-overlapping but differently-sized regions~\cite{fang2020near}. With this, more precision can be assigned to the range where a majority of values lie.
To model long-tail effects in data distribution, Biscaled-DNN uses two scale factors for quantization~\cite{biscaled_dnn}.
One scale factor is dedicated for increasing precision, and the other for increasing range.
ZeroQ sidetracks the need for a training dataset by engineering a synthetic one that matches the statistics of the batch normalization operation of each layer of the network~\cite{cai2020zeroq}.
This technique is considered another form of knowledge distillation.

Besides integer quantization, previous work proposes low-cost fixed-point and floating-point inspired data types for energy efficiency.
For example, Moons et al. proposes adaptive fixed-point quantization that trains a network for arbitrary fixed-point precision while minimizing energy~\cite{moons2017minimum}.
Flexpoint replaces 32-bit floating-point values with a block floating-point format that leverages shared exponents that can be dynamically adjusted to minimize overflow and maximize dynamic range~\cite{koster2017flexpoint}.
To avoid data loss from exponent sharing while improving energy efficiency, AdaptivFloat leverages a floating-point exponent bias based on absolute maximum value of the tensor to optimize the clipping of the tensor's dynamic range~\cite{tambealgorithm}.
Rouhani et al. explore the accuracy-cost tradeoffs of different variants of a block floating-point format in production-level cloud-scale inference~\cite{darvish2020pushing}.
Other work performs mixed-precision quantization to minimize bitwidth on a per-layer basis to adapt to each layer's sensitivity to precision~\cite{wu2018mixed,khoram2018adaptive}.

\section{Quantization Fundamentals}
\label{sec-quant}

Integer quantization maps high-precision floating-point weights and activations in a DNN to low-precision integer representations, typically with 8 or fewer bits.
For simplicity, in this paper we refer to the floating-point weights and activations collectively as real values, and the quantized low-precision weights and activations collectively as integer values.
We also focus on uniform integer quantization where the values are evenly distributed within the range of the integer format.
While non-uniform quantization such as logarithmic  quantization~\cite{miyashita2016convolutional} is also possible, the techniques proposed in this paper are orthogonal and can be applied to either form of quantization.

There are several considerations when deciding how to quantize real values into integers.
First, we must choose a range of real values to be represented so that any value out-of-range will be clipped. 
We may not necessarily want to choose the full range of presented real values, but rather clip outliers to improve the precision of quantized values within the range where most values reside.
Second, we need to select the number of bits available for our integer values.
With more integer bits, more discrete levels (integer values) are available to represent the same range of real values, resulting in smaller quantization error.

An $N$-bit signed two's complement integer quantization maps real values $x\in[x_{min},x_{max}]$ to values $x_q\in[-2^{N-1},2^{N-1}-1]$.
In general, a positive real scale factor $s$ is used to scale the value from the real range to the integer range, and a zero point $z$ represents the integer value corresponding to a real zero value.
Since the zero point complicates integer computation pipelines, efficient DNN accelerators typically apply symmetric scale-only quantization assuming $z=0$, $x_{min}=-x_{max}$, and $x_q\in[-2^{N-1}+1,2^{N-1}-1]$~\cite{wu2019low}.
If $\alpha$ denotes the absolute maximum real value that needs to be represented,
\begin{equation}
s=\frac{\alpha}{2^{N-1}-1}
\label{eq:scale_factor}
\end{equation}
\begin{equation}
x_q=clip\left(\left\lfloor\frac{x}{s}\right\rceil,-2^{N-1}+1,2^{N-1}-1\right)
\end{equation}
where $\left\lfloor\frac{x}{s}\right\rceil$ denotes rounding the scaled value to the nearest integer.
If $x$ is unsigned, $x_{min}=0$ and $x_q$ will be clipped to the integer range of $[0,2^{N-1}-1]$.
To avoid issues with vanishing gradient, quantized integer values $x_q$ are avoided during training.
Instead, simulated quantization using discrete real values is applied to simulate the effect of integer quantization~\cite{krishnamoorthi2018quantizing}.
Equation~\ref{eq:fake_quant_value} defines the simulated-quantized value $x_q^{s}$ as a real value from rescaling the integer value by the original scale factor.
\begin{equation}
x_q^{s}=s\cdot x_q
\label{eq:fake_quant_value}
\end{equation}

\begin{table*}[t]
\begin{center}
\small
 \begin{tabular}{c c c c c} 
 \hline
 Model          & Task           & Accuracy & Metric & Dataset\\ \hline\hline
 ResNet50 v1.5  & Image classification & 76.16    & Top1   & ImageNet 2012\\ \hline
 BERT-base      & Language model & 86.88    & F1     & SQuAD v1.1\\ \hline
 BERT-large     & Language model & 90.93    & F1     & SQuAD v1.1\\ \hline
\end{tabular}
\vspace{-3pt}
\caption{\BF{Overview of DNN models in this study}}
\vspace{3pt}
\label{tbl:model_overview}
\end{center}
\end{table*}

\begin{table*}[t]
\begin{center}
\small
 \begin{tabular}{c|c|ccccccc} 
 \hline
 Model                            & Bitwidths	& Max	     & Entropy	  & 99.9\%	  & 99.99\%	   & 99.999\%	& 99.9999\%  & MSE\\
 \hline\hline
 \multirow{4}{6em}{ResNet50} & Wt=3 Act=3U	& 0.18	     & 6.61	      & \bf{7.11} & 4.60	   & 1.54	    & 0.45       & 0.23 \\ 
                                  & Wt=4 Act=4U	& 60.31	     & \bf{70.76} & 70.46	  & 70.55	   & 68.71	    & 66.47      & 55.27 \\ 
                                  & Wt=6 Act=6U	& 75.43	     & 75.79	  & 75.24	  & \bf{75.80} & 75.73	    & 75.66      & 75.34 \\ 
                                  & Wt=8 Act=8U	& \bf{76.16} & 76.13	  & 75.37	  & 76.07	   & 76.11	    & 76.07      & 76.06 \\ \hline
 \multirow{3}{6em}{BERT-base} 
                              & Wt=4 Act=4	    & 2.23	     & 6.4	      & 5.66	  & \bf{6.91}  & 5.44	    & 0.06       & 6.82 \\ 
                              & Wt=6 Act=6	    & 7.31	     & 8.16	      & \bf{66.85}& 45.14	   & 14.16	    & 7.50       & 39.86 \\ 
                              & Wt=8 Act=8	    & 64.33	     & 63.7	      & 74.61	  & 80.63      & 78.95	    & 71.81      & \bf{81.25} \\ \hline
 \multirow{3}{6em}{BERT-large} 
 & Wt=4 Act=4	& 1.92	& \bf{6.81}	& 6.63	& 5.63	& 1.95	& 1.92 & 2.36 \\
 & Wt=6 Act=6	& 4.91	& 7.62	& \bf{32.29}	& 8.47	& 7.05	& 5.77 & 8.98 \\
 & Wt=8 Act=8	& 84.74	& 37.51	& 48.68	& 89.15	& \bf{89.41}	& 85.78 & 88.61 \\ \hline

\end{tabular}
\caption{\BF{DNN accuracy with per-channel scaling and static calibration:} Weight and activation bitwidths are specified under \TT{Bitwidths}. \TT{U} indicates unsigned values. Values are otherwise signed. \TT{Max}, \TT{Entropy}, and \TT{MSE} denote calibration using maximum absolute value, KL-divergence, and mean squared error, respectively. Percentages indicate the use of percentile calibration.}\vspace{5pt}
\label{tbl:baseline_results}
\end{center}
\end{table*}

Typically in a convolutional layer, a scale factor for weight or activation is determined for every layer of the network. 
Known as per-layer scaling, a single scale factor is used for each weight tensor (i.e., $K\times C\times R\times S$), and another scale factor is used for each activation tensor (i.e., $C\times H\times W$). 
To improve accuracy, multiple scale factors are determined for the weights of each layer. 
Known as per-channel scaling, a different scale factor is used for each output channel of a layer (i.e., $C\times R\times S$). 
We collectively refer to per-layer and per-channel scaling  as coarse-grained scaling.

Scale factors must be chosen carefully to best approximate a real distribution with a limited number of discrete values.
A calibration process is used to select the $\alpha$ used in Equation~\ref{eq:scale_factor} for quantizing weights and activations.
While $\alpha$ can be set to the maximum absolute value of the distribution (called max calibration), it is often beneficial to omit outlier values in favor of additional precision for inlier values.
For example, percentile calibration sets $\alpha$ to a specific fraction of $|x_{max}|$.
Entropy calibration, on the other hand, determines the $\alpha$ that minimizes the information loss between real and quantized distributions.
For weights, scale factors are determined using static calibration prior to inference.
For activations, scale factors can be determined using static calibration prior to inference or through dynamic calibration during inference.
Note that static calibration for activations requires samples of representative data to model the distribution of inputs that the network is likely to encounter during inference~\cite{wu2018mixed}.

While per-channel scaling achieves better accuracy than per-layer scaling, coarse-grained scaling methods generally lead to significant accuracy degradation for a range of quantized models. 
With PTQ but without QAT, we observe accuracy degradation in popular image recognition and language models after quantization, as indicated in Table~\ref{tbl:baseline_results}. 
Even for models where coarse-grained scaling can be competitive, careful calibration of the scale factor with the right calibration technique is required for good accuracy. 
As shown in Table~\ref{tbl:baseline_results}, the quality of calibration varies among different versions of the same network and across different networks.
We first focus on enabling state-of-the-art inference accuracy with PTQ before discussing {\em VS-Quant} for QAT.

\section{Per-vector Scaled Quantization}
\label{sec-pvsq}

\begin{table}[t]
\begin{center}
\small
 \begin{tabular}{c|c|c|c} 
 \hline
 Model & Bitwidths & Per-vector & Best Per-channel\\ \hline\hline
 \multirow{4}{3.8em}{ResNet50} & Wt=3 Act=3U & 69.78 & 7.97 \\
                             & Wt=4 Act=4U	& 75.28 & 70.76 \\ 
                             & Wt=6 Act=6U	& 76.00 & 75.80 \\ 
                             & Wt=8 Act=8U & 76.15 & 76.16 \\ \hline
 \multirow{4}{3.8em}{BERT-base} & Wt=3 Act=8	& 82.84 & 11.03\\
                              & Wt=4 Act=8	& 86.24 & 73.61 \\ 
                              & Wt=6 Act=8	& 86.66 & 80.18 \\ 
                              & Wt=8 Act=8	& 86.60 & 81.25 \\ \hline
\multirow{4}{3.8em}{BERT-large} & Wt=3 Act=8	& 89.56 & 8.71\\
                              & Wt=4 Act=8	& 90.64 & 83.18 \\ 
                              & Wt=6 Act=8	& 90.77 & 88.90 \\ 
                              & Wt=8 Act=8	& 90.80 & 89.41 \\ \hline
\end{tabular}
\vspace{-10pt}
\caption{\BF{PTQ accuracy of different DNN models with floating-point per-vector scale factors --} \TT{Best Per-Channel} indicates the best calibrated per-channel scaled quantized accuracy among all calibration methods in Table~\ref{tbl:baseline_results}.}
\label{tbl:fp_scaled_results}
\end{center}
\end{table}

We propose {\em VS-Quant}, per-vector scaled quantization, to mitigate the accuracy loss from quantization. 
Rather than computing a single scale factor over multiple dimensions of a tensor, {\em VS-Quant} applies a scale factor for each vector of elements within a single dimension of a tensor.
For a convolutional layer shown in Figure~\ref{fig:convolution}, per-vector scaling subdivides the input channel ($C$) dimension of the weight or activation tensor into $\lceil C/V\rceil$ vectors each with $V$ elements.
The number of vectors contained within a tensor depends on its shape and the designated vector size $V$.

In Table~\ref{tbl:fp_scaled_results}, we show that {\em VS-Quant} with static max calibration for weights and dynamic max calibration for activations has the potential to achieve significantly better accuracy with low bitwidths.
Compared to the floating-point baseline, per-vector scaled quantization achieves negligible accuracy drop at 6 bits and less than 1\% drop at 4 bits for ResNet50.
In comparison, per-channel scaled quantization requires at least 6-bit weights for less than 1\% drop.
Both BERT-base and BERT-large achieve close to full-precision accuracy with 4-bit weights, compared to per-channel scaled quantization which has difficulty reach the same level even with 8 bits.
Note that results are reported for PTQ where retraining is not required.

\subsection{Vector Size}

The quality of per-vector scaling depends on the vector size parameter.
At one extreme with $V=1$, each element would be individually quantized with its own scale factor and thus experience no loss in precision.
At the other extreme with $V=C$, elements in each $(R,S)$ in weight and $(H,W)$ in activation would share the same scale factor.
Table~\ref{tbl:vector_sizes} compares the accuracy of a 6-bit quantized ResNet50 with per-vector scaling for different vector sizes.  Accuracy decreases with increasing vector size because larger vectors have a higher probability of needing to represent a wider range of values.
The goal is to carefully select $V$ to minimize the required number of scale factors (maximize vector size) while maximizing the precision of the vector-scaled approximation and resulting network accuracy.

\begin{table}[t]
\begin{center}
\small
 \begin{tabular}{cccccccc} 
 \hline
V=1 & V=2 & V=4 & V=8 & V=16 & V=32 & V=64 \\\hline\hline
76.13 & 76.08 & 76.05 & 76.05 & 76.00 & 75.96 & 75.96\\\hline
\end{tabular}
\caption{\BF{Accuracy of 6-bit ResNet50 on ImageNet with {\em VS-Quant} for different vector sizes}}
\label{tbl:vector_sizes}
\end{center}
\end{table}

\subsection{Vector MAC}
\label{subsec-vector-mac}

In addition to better precision, the vector granularity also maps naturally to the vector unit of compute in typical DNN accelerators.
Because convolution and linear layers can be conveniently expressed as a collection of dot-products between an unrolled region of weights and an unrolled region of activations, vector-MAC units are the ubiquitous building blocks of many DNN processing architectures.
Equation~\ref{eq:dot_product} shows the dot-product $y(j)$ between the $j$th vector region of weights $w(j)(i),\;i\in [0,V-1]$ and the $j$th vector region of activations $a(j)(i),\;i\in[0,V-1]$.
\begin{equation}
y(j)=\sum_{i=0}^{V-1}(w(j)(i) \cdot a(j)(i))
\label{eq:dot_product}
\end{equation}
With {\em VS-Quant}, we compute a scale factor $s_w(j)$ for the $j$th weight vector and a scale factor $s_a(j)$ for the $j$th activation vector to scale the quantized integer weights $w_q(j)(i),\;i\in [0,V-1]$ and integer activations $a_q(j)(i),\;i\in[0,V-1]$.
Therefore, the dot-product in Equation~\ref{eq:dot_product} becomes the scaled dot-product in Equation~\ref{eq:scaled_dot_product}.
\begin{equation}
y_q^{s} (j)=\left (\sum_{i=0}^{V-1}(w_q (i) a_q (i))\right ) s_w (j)s_a (j)
\label{eq:scaled_dot_product}
\end{equation}
Note that the scale factors are factored out of each vector MAC, leading to a simple {\em VS-Quant} hardware implementation, as discussed in Section~\ref{sec-hw}.

\subsection{Calibration}

While it is orthogonal to per-vector scaling, calibration is still needed to determine the range of real values to be represented, which is parameterized by $\alpha$.
As with conventional scaling techniques, weight scale factors $s_w(j)$ can be determined statically based on the trained model.
Activation scale factors $s_a(j)$ can be computed statically with representative input samples or dynamically during inference.
Likewise, calibration methods including maximum absolute value, percentile, and entropy can still be applied.
However, because each vector only has a small number of elements, the distribution of a vector may lack enough samples to support more sophisticated calibration methods like percentile and entropy to determine a statistically useful $\alpha$.

\subsection{Two-Level Quantization}

The results in Table~\ref{tbl:fp_scaled_results} rely on floating-point scale factors per vector, which would lead to an inefficient hardware implementation.
To improve area and energy efficiency, we introduce a two-level scaling scheme that further applies integer quantization on the per-vector scale factors.
With this scheme, the per-vector scale factor $s$ in Equation~\ref{eq:fake_quant_value} is factored into the product of an integer per-vector scale factor $s_q$ and a floating-point coarse-grained scale factor $\gamma$, as shown in Equation~\ref{eq:vector_2level_scale}.
\begin{equation}
x_{q2}^{s}=s_q\cdot \gamma \cdot x_q
\label{eq:vector_2level_scale}
\end{equation}
Here $x_{q2}^{s}$ denotes the simulated-quantized values with two levels of scale factors.
With an integer scale factor per-vector, we need to store only a low-bitwidth integer alongside each vector of tensor elements, and we can complete all vector-wise computations with integer arithmetic.
With the two-level scaling technique, we push the more expensive floating-point scale factors to the coarser granularity by introducing the less expensive integer scale factors at the finer granularity to achieve a balance between accuracy and hardware efficiency.
Given N-bit integer weights or activations and M-bit integer per-vector scale factors, adding the M-bit scale factor alongside each $V$-element vector leads to a memory overhead of $M/(VN)$.
To give a perspective with $N=M=4$ and $V=16$, the storage overhead is 6.25\% which equates to an effective bitwidth of 4.25 bits.
Compared to coarse-grained scaling, two-level per-vector scaling requires scaling the dot-product by the product of the integer scale factors, which represents an extra $(2N+log(V))\times 2M$
multiplication for each vector dot-product.

Equations~\ref{subeq:vector_2level_max}-\ref{subseq:vector_2level_fake_quant_value} detail the algorithm for determining the scale factors when quantizing a real valued tensor $x$ to an N-bit signed integer in the two-level quantization scheme.
Index $i$ indicates each vector; index $j$ represents each element of a vector; and $k$ is the index along the coarse-grained dimension with different coarse-grained scale factors.
Assuming per-channel scale factors for the weight tensor of a convolutional layer, $k\in [0,K-1]$ while $i\in [0,\lceil C/V \rceil-1]$ and $j\in [0,V-1]$.

The algorithm first computes floating-point scale factors at a per-vector granularity.
Then it quantizes the per-vector scale factors by separating them into integer per-vector components and a floating-point per-channel component.
We specify the datatype of each tensor in Equation~\ref{eq:vector_2level} as $fp$ for floating-point and $int$ for integer.
\begin{subequations}\label{eq:vector_2level}
\begin{align}
x_{max}(k,i)_{fp}&=\max_{j}{|x(k,j,i)|}\label{subeq:vector_2level_max}\\
s(k,i)_{fp}&=\frac{x_{max}(k,i)}{(2^{N-1}-1)}\label{subeq:vector_2level_ratio}\\
x_q(k,j,i)_{int}&=\left\lfloor \frac{x(k,j,i)}{s(k,i)} \right\rceil\label{subeq:vector_2level_integer_value}\\
x_{q1}^{s}(k,j,i)_{fp}&=x_q(k,j,i)s(k,i)\label{subeq:vector_2level_single_fake_quant_value}\\
s_{max}(k)_{fp}&=\max_{i}{s(k,i)}\label{subseq:vector_2level_scale_max}\\
\gamma(k)_{fp}&=\frac{s_{max}(k)}{2^M-1}\\
s_q(k,i)_{int}&= \left\lfloor \frac{s(k,i)}{\gamma(k)} \right\rceil\label{subseq:vector_2level_integer_scale}\\
s_{q2}(k,i)_{fp}&={s_q(k,i)} \gamma(k)\label{subseq:vector_2level_scale_factor}\\
x_{q2}^{s}(k,j,i)_{fp}&=x_q(k,j,i)s_{q2}(k,i)\\
&=x_q(k,j,i)s_q(k,i)\gamma(k)\label{subseq:vector_2level_fake_quant_value}
\end{align}
\end{subequations}

To determine the per-vector scale factors, the algorithm computes the absolute maximum over the elements $j\in[0,V-1]$ of each vector $(k,i)$ in Equation~\ref{subeq:vector_2level_max} and then determines the floating-point per-vector scale factor that would scale the absolute maximum to the maximum representable N-bit signed integer.
This step is analogous to Equation~\ref{eq:scale_factor} but at a per-vector granularity.
Equation~\ref{subeq:vector_2level_integer_value} performs the actual per-vector scaling and rounds the resulting tensor values to integers which will be used in our integer dot-product unit.
Note that the scale factor here is per-vector for each $(k,i)$ but broadcast correspondingly to each element $(k,j,i)$ of the tensor.
At this point, we have everything we need if we were doing a single-level quantization with floating-point scale factors per-vector.
The single-level simulated-quantized value is expressed in Equation~\ref{subeq:vector_2level_single_fake_quant_value}.

To further quantize the scale factor, we repeat the quantization process of taking the absolute maximum, computing the ratio of real valued maximum to integer maximum, and scaling and rounding to integer on the single-level scale factor as shown in Equations~\ref{subseq:vector_2level_scale_max} to~\ref{subseq:vector_2level_integer_scale}.
Equation~\ref{subseq:vector_2level_scale_factor} shows the two-level scale factor as a composition of integer per-vector scale factor and floating-point per-channel scale factor.
The two-level simulated-quantized value is therefore represented as the product of the integer tensor values and the two levels of scale factors, as shown in Equation~\ref{subseq:vector_2level_fake_quant_value}.

\begin{table*}[t]
\begin{center}
\small
\setlength\tabcolsep{1.5mm}
\begin{tabular}{c|cccccc|c|c}
 \hline
    Bitwidths & S=3/4 & S=3/6 & S=4/4 & S=4/6 & S=6/4 & S=6/6 & S=fp32  & Best Per-channel \\ \hline\hline
    Wt=4 Act=3U & \cellcolor[rgb]{ .973,  .412,  .42}72.64 & \cellcolor[rgb]{ .976,  .624,  .631}73.51 & \cellcolor[rgb]{ .976,  .627,  .635}73.53 & \cellcolor[rgb]{ .98,  .824,  .831}74.33 & \cellcolor[rgb]{ .976,  .698,  .706}73.82 & \cellcolor[rgb]{ .984,  .91,  .922}74.69 & 74.71 & 67.20 \\
    Wt=4 Act=4U & \cellcolor[rgb]{ .976,  .592,  .604}73.39 & \cellcolor[rgb]{ .98,  .792,  .8}74.20 & \cellcolor[rgb]{ .98,  .831,  .839}74.36 & \cellcolor[rgb]{ .965,  .973,  .992}75.04 & \cellcolor[rgb]{ .984,  .882,  .894}74.58 & \cellcolor[rgb]{ .784,  .847,  .929}75.35 & 75.28 & 70.76 \\
    Wt=4 Act=6U & \cellcolor[rgb]{ .976,  .663,  .675}73.68 & \cellcolor[rgb]{ .984,  .851,  .863}74.45 & \cellcolor[rgb]{ .984,  .898,  .91}74.64 & \cellcolor[rgb]{ .847,  .89,  .953}75.25 & \cellcolor[rgb]{ .984,  .961,  .969}74.89 & \cellcolor[rgb]{ .761,  .827,  .922}75.40 & 75.40 & 72.20 \\
    Wt=4 Act=8U & \cellcolor[rgb]{ .976,  .659,  .667}73.65 & \cellcolor[rgb]{ .984,  .843,  .855}74.42 & \cellcolor[rgb]{ .984,  .902,  .914}74.66 & \cellcolor[rgb]{ .871,  .906,  .961}75.21 & \cellcolor[rgb]{ .984,  .945,  .957}74.83 & \cellcolor[rgb]{ .749,  .82,  .918}75.42 & 75.42 & 72.30 \\
    Wt=6 Act=3U & \cellcolor[rgb]{ .976,  .635,  .647}73.57 & \cellcolor[rgb]{ .98,  .831,  .839}74.36 & \cellcolor[rgb]{ .98,  .796,  .804}74.22 & \cellcolor[rgb]{ .984,  .984,  .996}74.99 & \cellcolor[rgb]{ .98,  .827,  .839}74.35 & \cellcolor[rgb]{ .804,  .859,  .937}75.32 & 75.23 & 71.52 \\
    Wt=6 Act=4U & \cellcolor[rgb]{ .98,  .804,  .816}74.26 & \cellcolor[rgb]{ .922,  .941,  .976}75.12 & \cellcolor[rgb]{ .984,  .976,  .988}74.95 & \cellcolor[rgb]{ .651,  .753,  .882}75.59 & \cellcolor[rgb]{ .906,  .933,  .973}75.14 & \cellcolor[rgb]{ .529,  .667,  .839}75.80 & 75.83 & 74.77 \\
    Wt=6 Act=6U & \cellcolor[rgb]{ .984,  .91,  .922}74.69 & \cellcolor[rgb]{ .914,  .937,  .976}75.13 & \cellcolor[rgb]{ .914,  .937,  .976}75.13 & \cellcolor[rgb]{ .561,  .69,  .851}75.74 & \cellcolor[rgb]{ .757,  .827,  .922}75.40 & \cellcolor[rgb]{ .435,  .6,  .808}75.96 & 76.00 & 75.80 \\
    Wt=6 Act=8U & \cellcolor[rgb]{ .984,  .875,  .886}74.55 & \cellcolor[rgb]{ .882,  .914,  .965}75.19 & \cellcolor[rgb]{ .882,  .914,  .965}75.19 & \cellcolor[rgb]{ .565,  .69,  .851}75.73 & \cellcolor[rgb]{ .753,  .824,  .918}75.41 & \cellcolor[rgb]{ .404,  .576,  .796}76.02 & 76.03 & 75.89 \\
    Wt=8 Act=3U & \cellcolor[rgb]{ .976,  .659,  .667}73.65 & \cellcolor[rgb]{ .984,  .855,  .867}74.47 & \cellcolor[rgb]{ .98,  .8,  .812}74.24 & \cellcolor[rgb]{ .918,  .937,  .976}75.13 & \cellcolor[rgb]{ .984,  .906,  .918}74.67 & \cellcolor[rgb]{ .788,  .847,  .929}75.35 & 75.56 & 71.98 \\
    Wt=8 Act=4U & \cellcolor[rgb]{ .984,  .859,  .871}74.48 & \cellcolor[rgb]{ .898,  .925,  .969}75.16 & \cellcolor[rgb]{ .941,  .957,  .984}75.08 & \cellcolor[rgb]{ .58,  .702,  .859}75.71 & \cellcolor[rgb]{ .871,  .906,  .961}75.21 & \cellcolor[rgb]{ .435,  .6,  .808}75.96 & 75.91 & 75.11 \\
    Wt=8 Act=6U & \cellcolor[rgb]{ .984,  .929,  .941}74.77 & \cellcolor[rgb]{ .804,  .859,  .937}75.32 & \cellcolor[rgb]{ .843,  .886,  .949}75.26 & \cellcolor[rgb]{ .494,  .639,  .827}75.86 & \cellcolor[rgb]{ .722,  .8,  .906}75.46 & \cellcolor[rgb]{ .404,  .58,  .796}76.01 & 76.17 & 76.01 \\
    Wt=8 Act=8U & \cellcolor[rgb]{ .984,  .89,  .902}74.61 & \cellcolor[rgb]{ .8,  .855,  .933}75.33 & \cellcolor[rgb]{ .906,  .929,  .973}75.15 & \cellcolor[rgb]{ .502,  .647,  .831}75.85 & \cellcolor[rgb]{ .718,  .8,  .906}75.47 & \cellcolor[rgb]{ .353,  .541,  .776}76.10 & 76.15 & 76.16 \\
    \hline
\end{tabular}
\end{center}
\vspace{-10pt}
\caption{\BF{ResNet50 on ImageNet with integer per-vector scale factors} }
\label{tbl:int_scaled_resnet50_results}
\vspace{5pt}

\begin{center}
\small
 \begin{tabular}{c|cccc|c|c|c} 
 \hline
Bitwidths & S=4/8 & S=4/10 & S=6/8 & S=6/10 & S=fp16 & S=fp32 & Best Per-channel \\ \hline\hline
    Wt=3 Act=8 & \cellcolor[rgb]{ .973,  .412,  .42}76.28 & \cellcolor[rgb]{ .98,  .776,  .788}81.84 & \cellcolor[rgb]{ .973,  .553,  .561}78.44 & \cellcolor[rgb]{ .98,  .843,  .851}82.81 & \cellcolor[rgb]{ .984,  .847,  .859}82.90 & 82.93 & 11.03 \\
    Wt=4 Act=8 & \cellcolor[rgb]{ .984,  .847,  .855}82.87 & \cellcolor[rgb]{ .659,  .757,  .886}85.91 & \cellcolor[rgb]{ .984,  .878,  .89}83.39 & \cellcolor[rgb]{ .498,  .643,  .827}86.35 & \cellcolor[rgb]{ .502,  .647,  .831}86.34 & 86.33 & 73.61 \\
    Wt=6 Act=8 & \cellcolor[rgb]{ .984,  .886,  .898}83.47 & \cellcolor[rgb]{ .569,  .694,  .855}86.16 & \cellcolor[rgb]{ .984,  .894,  .906}83.63 & \cellcolor[rgb]{ .427,  .596,  .804}86.54 & \cellcolor[rgb]{ .416,  .588,  .8}86.57 & 86.61 & 80.18 \\
    Wt=8 Act=8 & \cellcolor[rgb]{ .984,  .89,  .902}83.53 & \cellcolor[rgb]{ .506,  .647,  .831}86.33 & \cellcolor[rgb]{ .984,  .902,  .914}83.75 & \cellcolor[rgb]{ .408,  .58,  .796}86.59 & \cellcolor[rgb]{ .353,  .541,  .776}86.74 & 86.74 & 81.25 \\
\hline
\end{tabular}
\end{center}
\vspace{-10pt}
\caption{\BF{BERT-base on SQuAD with integer per-vector scale factors}}
\label{tbl:int_scaled_bert_base_results}
\vspace{5pt}

\begin{center}
\small
 \begin{tabular}{c|cccc|c|c|c} 
 \hline
Bitwidths & S=4/8 & S=4/10 & S=6/8 & S=6/10 & S=fp16 & S=fp32 & Best Per-channel \\ \hline\hline
    Wt=3 Act=6 & \cellcolor[rgb]{ .973,  .412,  .42}83.17 & \cellcolor[rgb]{ .98,  .714,  .722}86.24 & \cellcolor[rgb]{ .973,  .522,  .529}84.3 & \cellcolor[rgb]{ .98,  .78,  .792}86.92 & \cellcolor[rgb]{ .98,  .792,  .8}87.03 & 87.13 & 6.88 \\
    Wt=3 Act=8 & \cellcolor[rgb]{ .98,  .773,  .78}86.82 & \cellcolor[rgb]{ .984,  .973,  .984}88.86 & \cellcolor[rgb]{ .984,  .851,  .863}87.65 & \cellcolor[rgb]{ .816,  .867,  .941}89.5 & \cellcolor[rgb]{ .769,  .835,  .925}89.63 & 89.63 & 8.71 \\
    Wt=4 Act=6 & \cellcolor[rgb]{ .98,  .796,  .808}87.08 & \cellcolor[rgb]{ .984,  .894,  .906}88.08 & \cellcolor[rgb]{ .984,  .855,  .863}87.66 & \cellcolor[rgb]{ .984,  .937,  .949}88.52 & \cellcolor[rgb]{ .984,  .969,  .98}88.84 & 88.84 & 10.06 \\
    Wt=4 Act=8 & \cellcolor[rgb]{ .8,  .859,  .937}89.54 & \cellcolor[rgb]{ .533,  .667,  .839}90.31 & \cellcolor[rgb]{ .725,  .804,  .91}89.76 & \cellcolor[rgb]{ .451,  .612,  .812}90.54 & \cellcolor[rgb]{ .369,  .553,  .784}90.78 & 90.78 & 83.18 \\
    Wt=6 Act=6 & \cellcolor[rgb]{ .98,  .824,  .831}87.34 & \cellcolor[rgb]{ .984,  .925,  .937}88.39 & \cellcolor[rgb]{ .984,  .863,  .875}87.76 & \cellcolor[rgb]{ .988,  .988,  1}89.01 & \cellcolor[rgb]{ .937,  .953,  .984}89.15 & 89.2  & 32.29 \\
    Wt=6 Act=8 & \cellcolor[rgb]{ .765,  .831,  .922}89.65 & \cellcolor[rgb]{ .478,  .627,  .82}90.47 & \cellcolor[rgb]{ .706,  .788,  .902}89.82 & \cellcolor[rgb]{ .42,  .588,  .8}90.63 & \cellcolor[rgb]{ .353,  .541,  .776}90.82 & 90.81 & 88.9 \\
    Wt=8 Act=6 & \cellcolor[rgb]{ .98,  .827,  .839}87.4 & \cellcolor[rgb]{ .984,  .953,  .965}88.68 & \cellcolor[rgb]{ .984,  .859,  .871}87.72 & \cellcolor[rgb]{ .984,  .984,  1}89.02 & \cellcolor[rgb]{ .937,  .953,  .984}89.15 & 89.13 & 40.5 \\
    Wt=8 Act=8 & \cellcolor[rgb]{ .698,  .784,  .898}89.84 & \cellcolor[rgb]{ .478,  .627,  .82}90.47 & \cellcolor[rgb]{ .6,  .714,  .863}90.12 & \cellcolor[rgb]{ .412,  .584,  .8}90.66 & \cellcolor[rgb]{ .369,  .553,  .784}90.78 & 90.78 & 89.41 \\
\hline
\end{tabular}
\end{center}
\vspace{-10pt}
\caption{\BF{BERT-large on SQuAD with integer per-vector scale factors}}
\label{tbl:int_scaled_bert_large_results}
\bigskip
\flushleft\vspace{-6pt}
\IT{Tables~\ref{tbl:int_scaled_resnet50_results}-\ref{tbl:int_scaled_bert_large_results}.} \BF{Accuracy of different networks when applying integer per-vector scale factors --} Accuracy numbers are color-coded from highest (dark blue) to lowest acceptable (dark red). \TT{S=Sw/Sa} indicates \TT{Sw}-bit unsigned per-vector weight scale factors and \TT{Sa}-bit unsigned per-vector activation scale factors. \TT{S=fp16} and \TT{S=fp32} indicate single-level fp16 and fp32 per-vector scale factors.\vspace{8pt}
\end{table*}

Using two-level quantization for calibrating scale factors, DNN inference accuracy with PTQ across a range of weight, activation, and scale factor bitwidths is shown in Tables~\ref{tbl:int_scaled_resnet50_results}, ~\ref{tbl:int_scaled_bert_base_results}, and~\ref{tbl:int_scaled_bert_large_results}.  We compare the accuracy of {\em VS-Quant} with two-level scaling using low-bitwidth integer and fp16 scale factors to {\em VS-Quant} with fp32 scale factors and per-channel scaling (similar to Table~\ref{tbl:fp_scaled_results}).  Compared to per-vector scaling, we consistently observe significantly lower accuracy loss with {\em VS-Quant} across all three DNNs, particularly at low weight and activation bitwidths.  For example, {\em VS-Quant} with 3-bit weights and 8-bit activations achieves over 89\% accuracy for BERT-large on SQuAD while the best per-channel calibrated quantization only achieves 8.7\% accuracy.

The two-level quantization algorithm in Equation~\ref{eq:vector_2level} is merely one of several ways to determine the two levels of scale factors.
For example, instead of first computing the single-level per-vector scale factor and then breaking it down into the product of two levels of scale factors, we can do it one level at a time by first computing the per-channel scale factor and then back-calculating the per-vector scale factor.
While this approach provides a larger space to explore the integer values and integer scale factors, it requires computing the absolute maximum over a larger tensor as opposed to just a vector.
This is much more expensive to implement in hardware if scaling activations dynamically during inference.
However, it could be acceptable for scaling weights statically before inference.

\section{Hardware Implementation}
\label{sec-hw}

\begin{figure*}[tbh]
\centering
\includegraphics[trim=0pt 120pt 30pt 0pt, clip,width=0.99\textwidth]{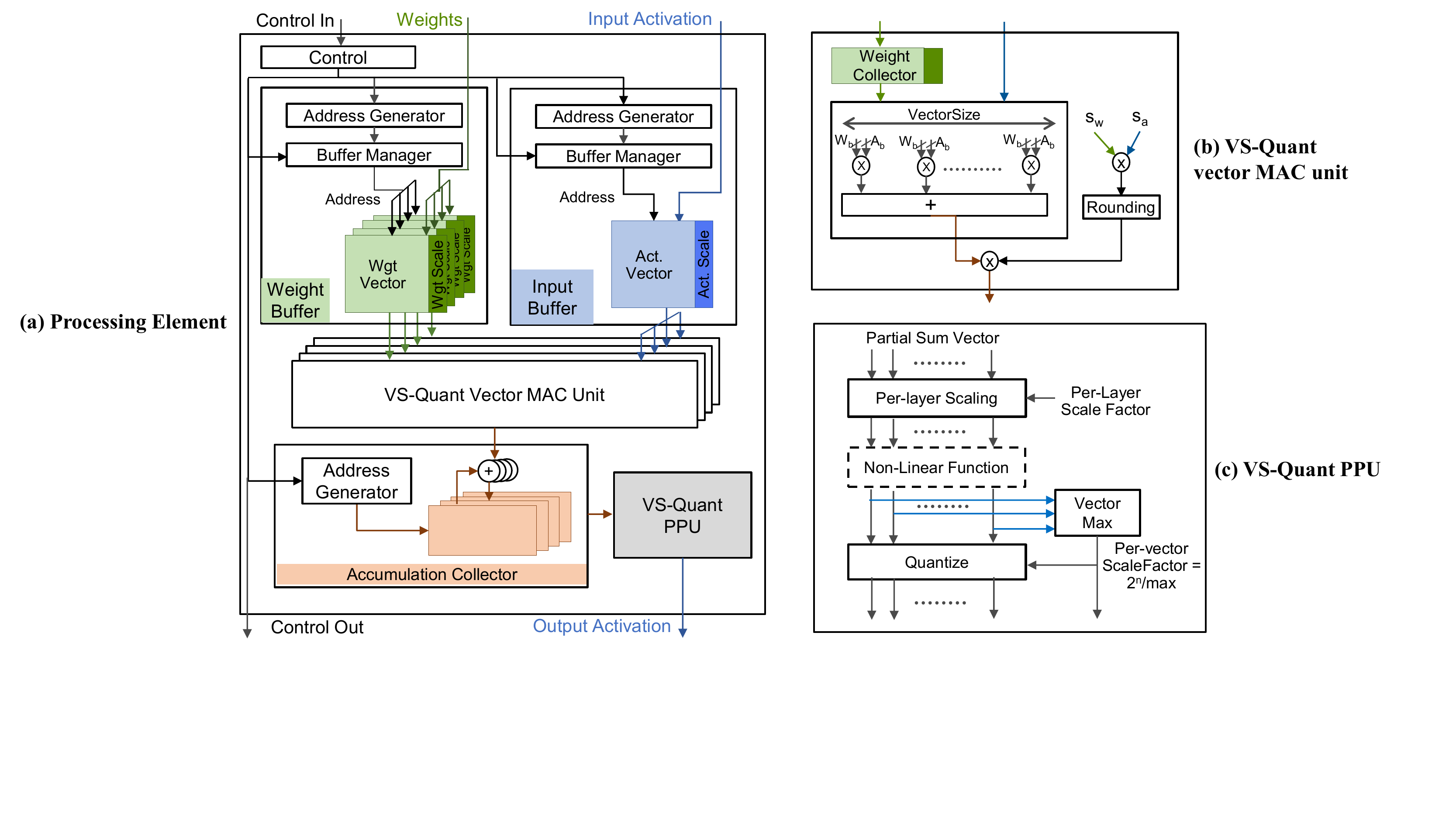}
\vspace{-0.10in}
\caption{\BF{Hardware diagram} --- DNN accelerator with per-vector scaling support.}\vspace{10pt}
\label{fig:pvs_hardware}
\vspace{3pt}
\end{figure*}

To evaluate the hardware efficiency of {\em VS-Quant}, we extended a previous optimized DNN accelerator~\cite{venkatesan2019magnet} by adding per-vector scaling support. Figure~\ref{fig:pvs_hardware}(a) shows the micro-architecture of a processing element (PE) in the accelerator, which is responsible for the dot-product computation listed in Equations~\ref{eq:dot_product} and~\ref{eq:scaled_dot_product}. The PE consists of a set of {\em VS-Quant} vector MAC units, a weight buffer, an input activation buffer, an accumulation collector, a {\em VS-Quant} post-processing unit, and control logic. 

Each {\em VS-Quant} vector MAC unit, shown in Figure~\ref{fig:pvs_hardware}(b), performs a $V$-element dot-product between the corresponding weight and activation data. In parallel, the product of the per-vector weight scale factor $s_w$ and activation scale factor $s_a$ is computed and rounded to the desired precision. The two outputs are then multiplied to compute a scaled partial sum output. 
Each entry of the weight buffer stores a weight vector along with corresponding per-vector scale factor. Similarly, the input activation buffer stores an activation vector and a per-vector scale factor in each row. The accumulation collector stores partial sum values from all the vector MAC computations and temporally accumulates them across multiple cycles in an integer format. For N-bit weights and activations along with M-bit weight and activation scale factors, we have $N\times N\rightarrow 2N$-bit products that are accumulated over the vector size $V$, resulting in $2N+log_2{V}$ wide dot-product outputs.  
The dot-product results are multiplied with the product of the M-bit weight and activation scale factors to produce $2N+log_2{V} + 2M$ wide partial sums. 
For improved energy efficiency, the vector MAC unit can optionally round the product of the scale factors to fewer than $2M$ bits before multiplying with the dot-product result.
Finally, the accumulation collectors are designed with appropriate widths to avoid overflow. Taken together, the PE achieves efficient data reuse across all three data types: (i) each input activation vector is shared spatially across multiple vector MAC units; (ii) weight vectors are reused temporally across multiple cycles using a weight collector; (iii) partial sums are reused spatially inside the vector MAC unit and temporally in the accumulation collector. 

For post-processing, the output of the accumulation collector is fed to a post-processing unit (PPU). To implement dynamic calibration for the scale factors of the activations, we perform the required calibration operations in the {\em VS-Quant} PPU and convert the higher-precision output activations back to N-bit vector elements with per-vector scale factors for the next layer. Figure~\ref{fig:pvs_hardware}(c) shows the block diagram of the {\em VS-Quant} PPU that performs the calibrate-and-quantize operations.
As a post-processing step following the completion of a layer of computation, we leverage a vector max unit to implement Equation~\ref{subeq:vector_2level_max} to compute the absolute maximum of each vector of elements.
Then a reciprocal unit and shifter implement Equation~\ref{subeq:vector_2level_ratio} to compute the ratios of absolute maximums of the vector to the maximum representable value of an N-bit integer value. 
The computed ratios are the scale factors used to quantize the output activations and convert them to {\em VS-Quant} format for computation of the next layer.

To quantify the area and energy impact of supporting {\em VS-Quant} in hardware, we also consider a baseline PE architecture for comparison without the scale factor related multipliers in the vector MAC unit and without the scale factor overheads in the weight and activation buffers. In this case, each vector MAC unit simply performs a $V$-wide dot-product and produces a partial sum of width $2N+log_2{V}$ for N-bit weights and activations.  Per-channel scaling is performed in the baseline design PPU.  

We evaluate the impact on energy per operation of {\em VS-Quant} compared to the baseline design using the MAGNet DNN generator and exploration infrastructure~\cite{venkatesan2019magnet}.  MAGNet's published 8-bit configuration achieved 2.1 tera-operations/sec/mm\textsuperscript{2} (TOPS/mm\textsuperscript{2}) and 69 fJ/operation (14.5 TOPS/Watt) in a 16nm FinFET technology node. We normalize all subsequent energy and area numbers in this paper to a similar baseline design with 8-bit weights and activations. The design tools shown in Table~\ref{tbl:experimental_setup} are used to implement the hardware and measure area and power in a sub-16nm process technology. 

\begin{table}[t]
\begin{center}
\small
 \begin{tabular}{c|c} 
 \hline\hline
\multicolumn{2}{c}{Design tools} \\ \hline\hline
HLS Compiler & Mentor Graphics Catapult HLS\\ \hline
Verilog simulator & Synopsys VCS\\ \hline
Logic synthesis & Synopsys Design Compiler Graphical\\ \hline
Place-and-route & Synopsys ICC2 \\ \hline
Power analysis & Synopsys PT-PX \\ \hline
\hline
\multicolumn{2}{c}{Design space} \\ \hline\hline
Vector size & 16 \\ \hline
Weight/activation \\precision & 3-bit, 4-bit, 6-bit, 8-bit \\ \hline
Weight/activation \\scale precision & 3-bit, 4-bit, 6-bit, 8-bit, 10-bit \\ \hline
Scaling granularity & POC, PVAO, PVWO, PVAW \\ \hline
\hline
\end{tabular}
\caption{\BF{Experimental setup --} \TT{POC} = per-channel, \TT{PVAO} = per-vector on activations only, \TT{PVWO} = per-vector on weights only, and \TT{PVAW} = per-vector on both weights and activations.}
\label{tbl:experimental_setup}
\vspace{3pt}
\end{center}
\end{table}

\begin{figure}[t]
\centering
\includegraphics[trim=0pt 50pt 0pt 0pt, width=0.95\columnwidth]{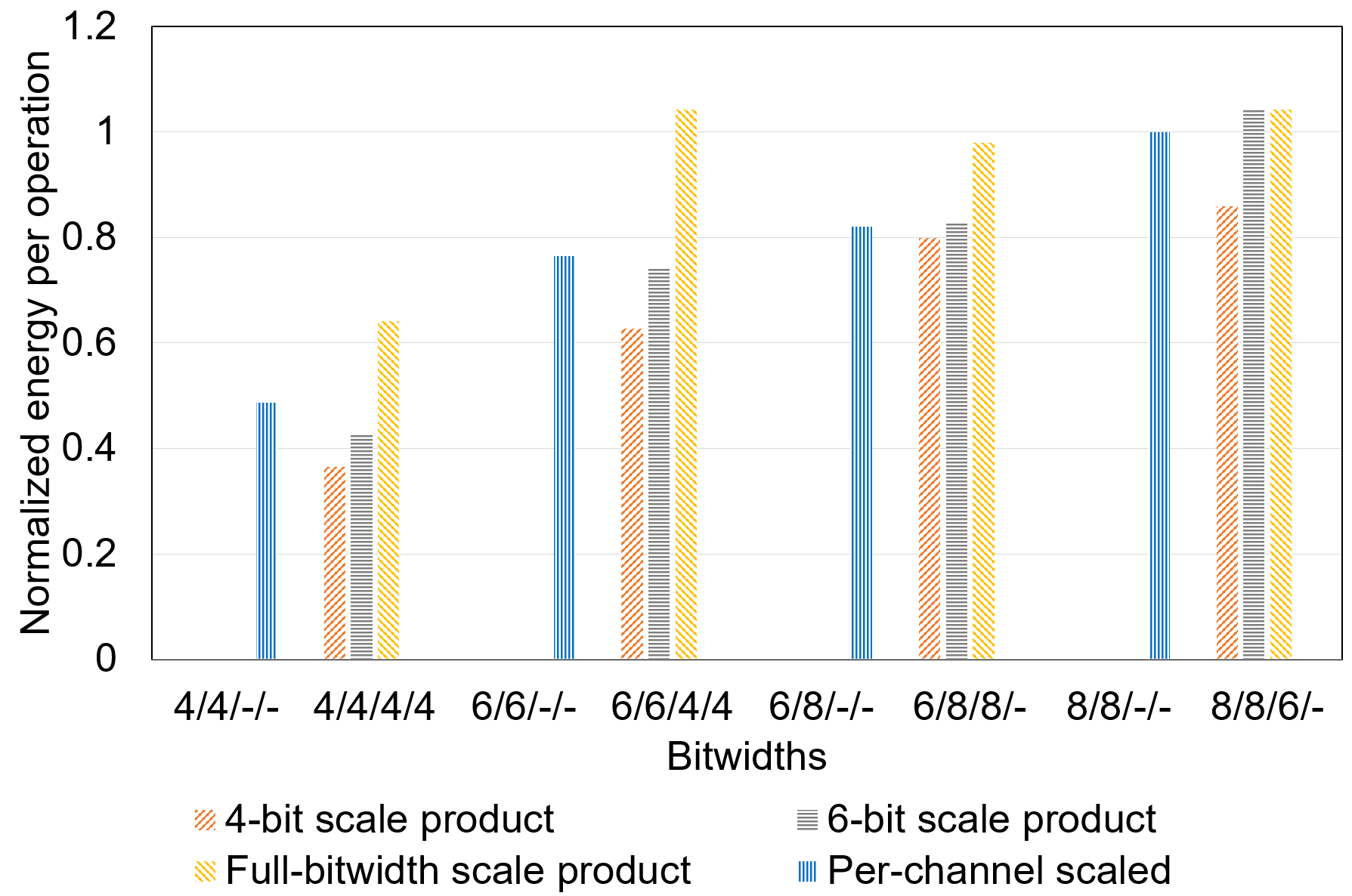}
\caption{\BF{Effect of scale product bitwidth on energy --} For this and subsequent figures, \TT{W/A/ws/as} indicates weight, activation, weight scale, and activation scale bitwidths. Dashes indicate per-channel/per-layer scaling for weights/activations, respectively. }
\label{fig:reduced_scale_product_space}
\vspace{3pt}
\end{figure}

Figure~\ref{fig:reduced_scale_product_space} shows the average energy per operation across a range of hardware configurations.  In this and all subsequent plots, we use \TT{W/A/ws/as} to denote each configuration, where \TT{W} stands for weight bitwidth, \TT{A} for activation bitwidth, \TT{ws} for weight scale bitwidth, and \TT{as} for activation scale bitwidth.
\TT{-} indicates use of per-channel scaling.
Energy is normalized to that of the 8/8/-/- configuration.
The blue bars for the per-channel scaled configurations (4/4/-/-, 6/6/-/-, 6/8/-/-, 8/8/-/-) show that quantization can achieve up to 2x energy savings over an 8-bit baseline. 
When the {\em VS-Quant} hardware is introduced and the scale factor product ($s_w \times s_a$) in Figure~\ref{fig:pvs_hardware}(b) is kept at full-bitwidth precision (i.e., no rounding), the yellow bars for the 4/4/4/4 and 6/6/4/4 configurations show modest energy overheads at 4-bit and 6-bit weight and activation precisions over corresponding per-channel scaled configurations due to additional multipliers for scaling and wider accumulation widths. 
When the scale factor product is rounded to an intermediate size of 4 bits or 6 bits, the energy overheads of adding {\em VS-Quant} support to the hardware can be substantially reduced, as demonstrated by the orange and grey bars. 
In fact, scale factor rounding truncates many small values and converts them to zero, thereby providing opportunities for data gating of costly accumulation operations. 
As a result, the configurations with scale product rounding can achieve lower energy consumption compared to even the per-channel scaled configurations.
The 8/8/6/- configuration shows the same energy for 6-bit scale and full-bitwidth scale product because full-bitwidth is exactly 6 bits in this case because of its 6-bit per-vector weight scale factor and no per-vector activation scale factor. 

\section{Design Space Exploration}
\label{sec-dse}

To better understand the accuracy, energy, and area tradeoffs enabled by {\em VS-Quant}, we combine the energy and area results from our DNN inference accelerator with accuracy results from real networks using a Pytorch-based PTQ library~\cite{wu2020integer}.  
Table~\ref{tbl:experimental_setup} details the design tools used and parameters explored in our full evaluation.  In this section, we limit DNN accelerator configurations to those without intermediate rounding and use full-bitwidth scale factor products (yellow bars from Figure~\ref{fig:reduced_scale_product_space}).

\begin{figure}[t]
\centering
\includegraphics[trim=0pt 10pt 0pt 50pt, clip, width=\columnwidth]{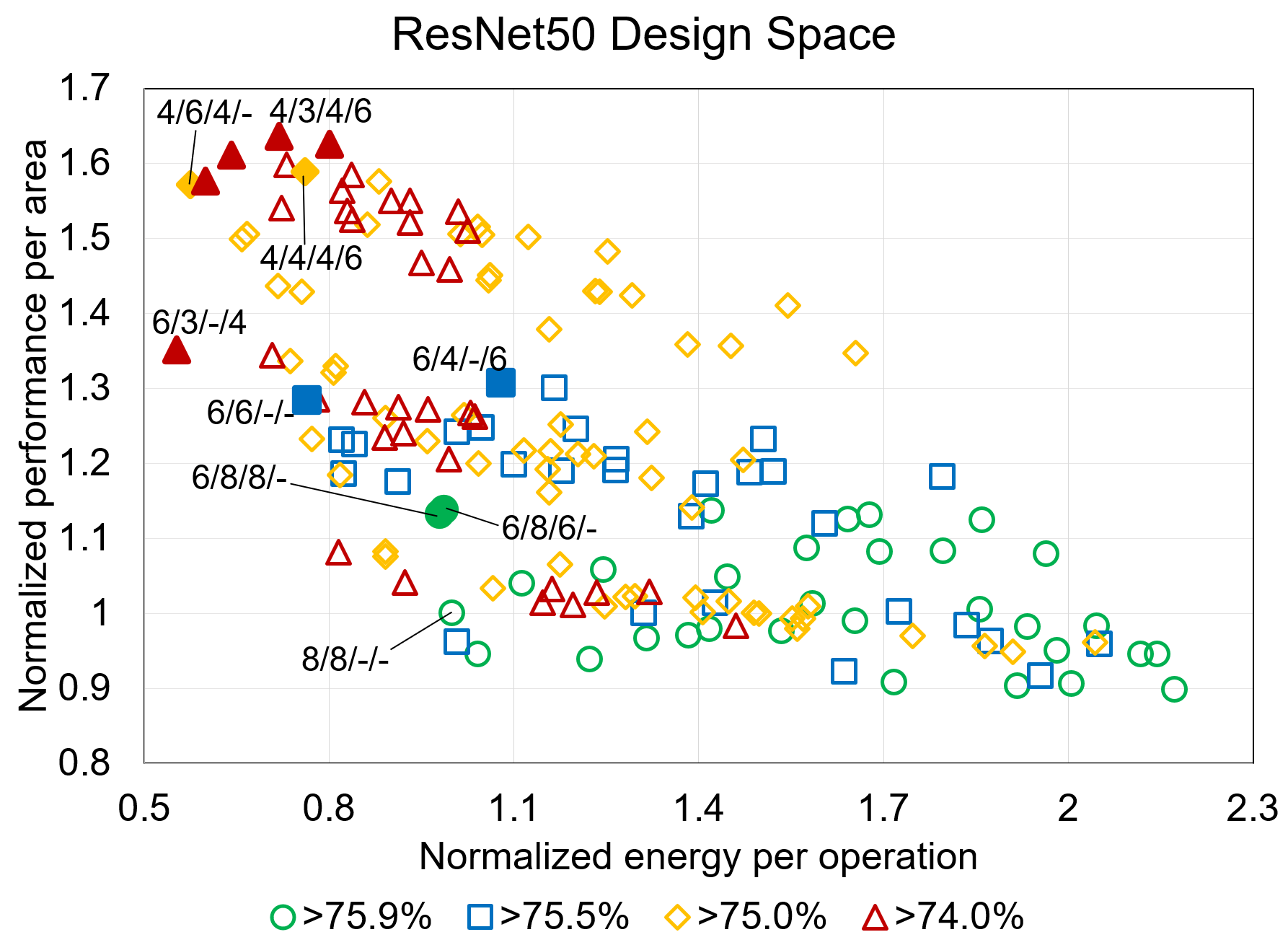}
\vspace{-18pt}
\caption{\BF{ResNet50 design space}}
\label{fig:resnet_design_space}
\vspace{5pt}
\end{figure}

\begin{figure*}[t]
\centering
\begin{minipage}[b]{0.48\linewidth}
\includegraphics[trim=0pt 10pt 0pt 50pt, clip, width=0.95\columnwidth]{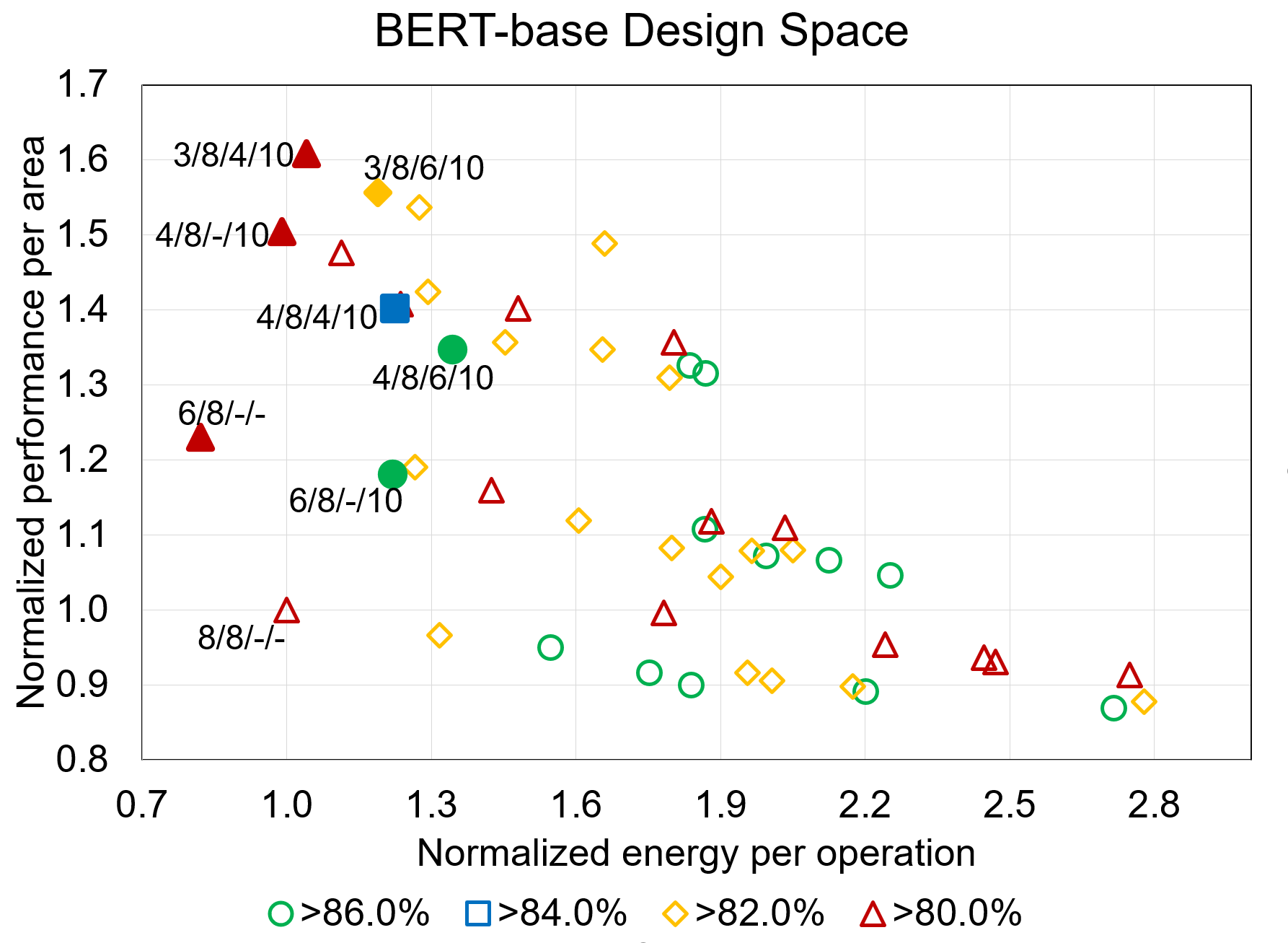}
\vspace{-10pt}
\caption{\BF{BERT-base design space} }
\label{fig:bert_base_design_space}
\end{minipage}
\hspace{3pt}
\begin{minipage}[b]{0.48\linewidth}
\includegraphics[trim=0pt 10pt 0pt 50pt, clip, width=0.95\columnwidth]{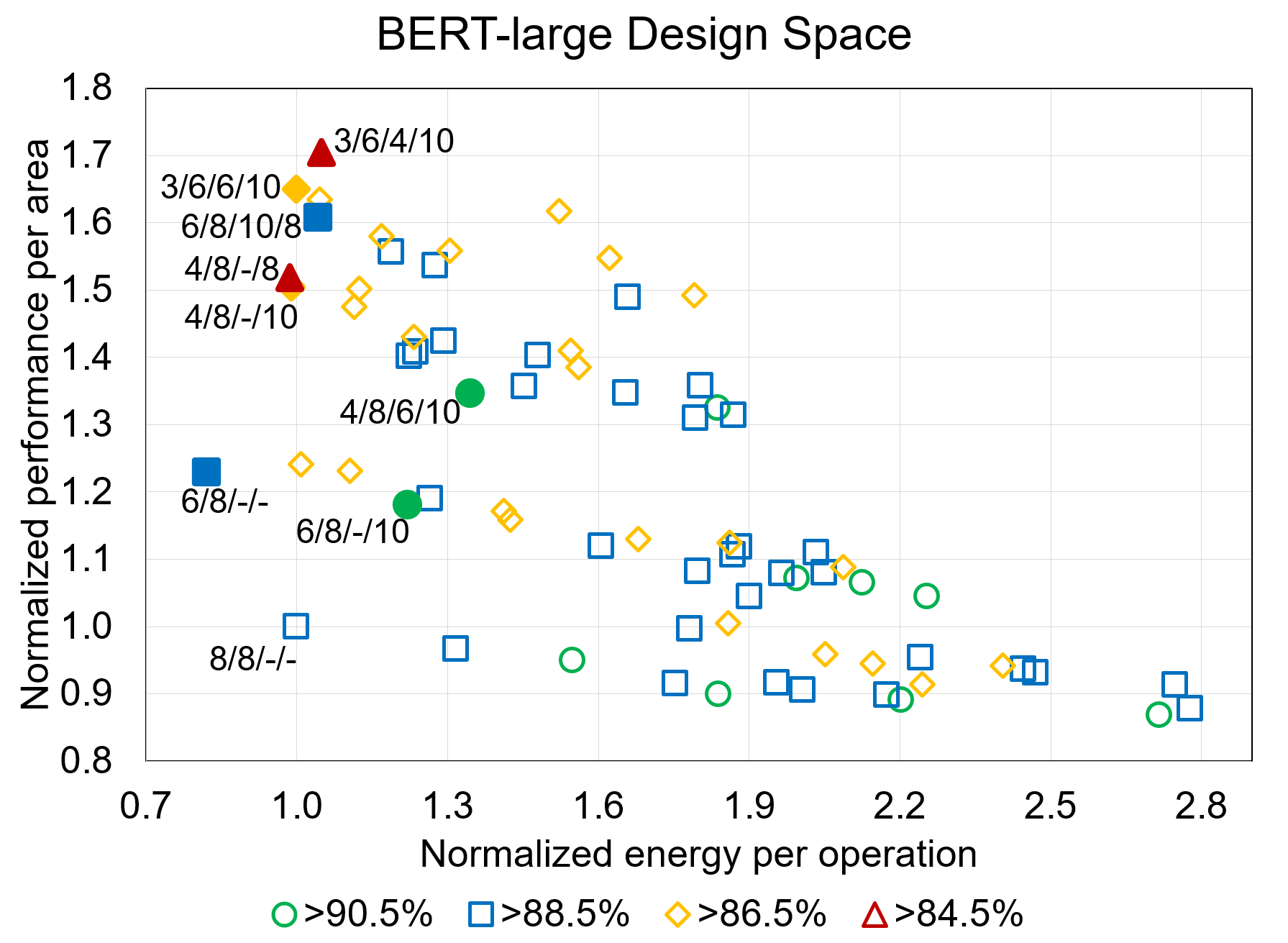}
\vspace{-10pt}
\caption{\BF{BERT-large design space} }
\label{fig:bert_large_design_space}
\end{minipage}
\vspace{6pt}
\end{figure*}

Figures~\ref{fig:resnet_design_space}, ~\ref{fig:bert_base_design_space},  and~\ref{fig:bert_large_design_space} present the design spaces of ResNet50, BERT-base, and BERT-large, respectively, for various bitwidth configurations of our DNN accelerator hardware.
Results are shown as a tradeoff among energy efficiency (x-axis), area efficiency as performance per unit area (y-axis), and inference accuracy (color/shape).  Since all configurations run with the same throughput (operations per cycle), performance is identical and only the VLSI energy and area costs vary.
Each point in the plot reports metrics for a synthesized hardware instance selected from the set of precision parameter options in Table~\ref{tbl:experimental_setup}, normalized to our baseline design (8/8/-/- configuration).
Energy results are averaged over layers of the networks, weighted by the number of operations in each layer.
For each network, we decide the acceptable amount of accuracy loss against the full-precision baseline and only visualize those design points that are within the acceptable accuracy range.
As indicated in the legends in Figures~\ref{fig:resnet_design_space}, ~\ref{fig:bert_base_design_space}, and ~\ref{fig:bert_large_design_space}, we plot only ResNet50, BERT-base, and BERT-large design points that have an accuracy above 74.0\%, 80.0\%, and 84.5\%, respectively.
We then subdivide the acceptable range into finer accuracy ranges (four colors/shapes) to help visualize the achieved accuracy on top of the area-energy space. 
For design points of the same color/shape (within the same accuracy range), the upper left of the plot is optimal with the lowest energy per operation and highest performance per area.  
Solid points indicate Pareto-optimal area or energy efficiency for their color/shape (accuracy range) whereas hollow points are not optimal.
Overall, {\em VS-Quant} provides a much more expansive space of design tradeoffs than baseline 4-bit, 6-bit, and 8-bit datapaths, which we discuss in detail for each network below.

For ResNet50 results (Figure~\ref{fig:resnet_design_space}), the baseline 8/8/-/- already has minimal accuracy loss compared to the floating-point reference, so limited accuracy gains are available from {\em VS-Quant}. 
However, the green/circle 6/8/6/- {\em VS-Quant} point (6-bit weights, 8-bit activations, and 6-bit per-vector scale factors for weights) provides 12\% smaller area at similar accuracy and similar energy.
When moving to 4-bit and 6-bit representations,  {\em VS-Quant} provides even more energy and area reductions in the 74.5\% to 75.5\% accuracy range.  
For example, in the $>$75.0\% accuracy range (yellow/rhombus points), the 4/6/4/- design point achieves 43\% less energy and 36\% smaller area than the baseline design.
When moving to even lower accuracy in the $>$74.0\% range (red/triangle points), even lower energy can be found at the 6/3/-/4 point or smaller area at the 4/3/4/6 point.  
In prior work, limiting accuracy loss to 1-2\% with 4-to-6-bit integer representations had only been possible with QAT.

Figures~\ref{fig:bert_base_design_space} and ~\ref{fig:bert_large_design_space} highlight the best energy-efficiency and area achieved for BERT-base and BERT-large similarly at different accuracy targets.
For both BERT models, {\em VS-Quant} is observed to be the most competitive across multiple accuracy targets, requiring very few bits for representing weights.
In particular, a 4/8/6/10 configuration (4-bit weights, 8-bit activations, 6-bit per-vector weight scale factors, and 10-bit per-vector activation scale factors) for either model can achieve an accuracy target close to that of the full-precision baseline.
This accuracy is not attainable even with our baseline design (8-bit per-channel scaled quantization) according to Table~\ref{tbl:baseline_results}.
Alternatively, we can also save some energy at the cost of a slight area increase with a 6/8/-/10 configuration while maintaining close to full-precision accuracy.
Although this configuration requires a higher weight bitwidth than the previous configuration, it reduces energy by avoiding per-vector scaling on the weights.
This tradeoff suggests a combined effect between the value bitwidth and scale factor bitwidth, which together present an effective bitwidth for the particular configuration.
If we relax our accuracy requirement to at least 82.0\% for BERT-base and 86.5\% for BERT-large, we can further decrease area and energy by dropping weight precision to only 3 bits.
Based on the design points, the only BERT-large configuration where it makes sense to implement per-channel scaled quantization is the 6/8/-/- configuration targeting around 1\% accuracy loss, although this configuration trades off significant area to attain the lowest energy in that accuracy range.
Furthermore, if the same 6/8/-/- hardware configuration was chosen for BERT-base, it would lead to a large 6\% accuracy loss. In comparison, optimal {\em VS-Quant} hardware configurations such as 4/8/6/10 achieve great accuracy on both BERT-base and BERT-large.

\begin{figure}[t]
\centering
\includegraphics[trim=0pt 10pt 0pt 50pt,clip, width=0.9\columnwidth]{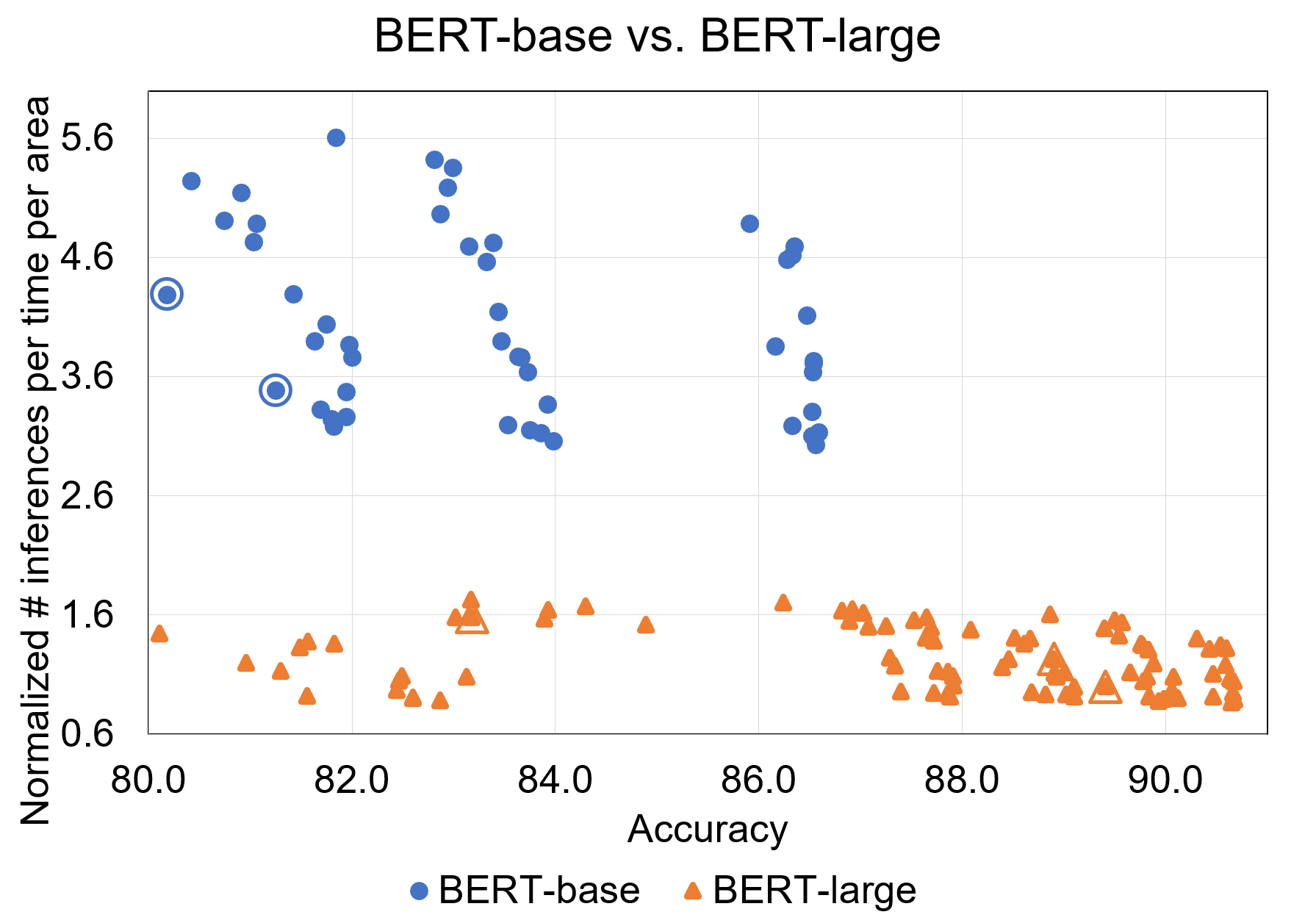}
\vspace{-10pt}
\caption{\BF{Accuracy and area tradeoff for BERT models of different sizes --} Per-channel design points are outlined. }
\label{fig:bert_performance_comparison}
\vspace{-3pt}
\end{figure}

We further study how the size of a network affects its accuracy, energy, and area tradeoff by comparing the design points of BERT-base against those of BERT-large.
As shown in Figure~\ref{fig:bert_performance_comparison}, for example, BERT-large is the only choice if the accuracy target is beyond the best that BERT-base is able to achieve.
Below that threshold, we should always select BERT-base because it is consistently more area-efficient than BERT-large.
This suggests that one should configure the size of the model based on the desired accuracy target to realize the best hardware efficiency.

\section{Quantization-aware Retraining}
\label{sec-discussions}

While we are able to leverage per-vector scaled PTQ to maintain reasonable accuracy down to 3 bits in some cases, accuracy loss is inevitable at low precisions when compared to a full-precision baseline when QAT is not applied.
The loss can be substantial if an inferior combination of weight and activation precisions is used.
For example, BERT generally requires 8-bit precision for activations to get reasonable accuracy even with {\em VS-Quant}.
In addition, many practical deployment scenarios may not have QAT as an option due to lack of access to full training datasets or limits on compute time and tuning effort.  
However, there are cases in which we can finetune a pretrained model with quantization for only a limited number of iterations to adapt the weights and activations to the quantized model~\cite{mckinstry2018discovering}.

{\em VS-Quant} is not limited to PTQ and can also be applied to QAT to achieve even higher accuracy for a given set of bitwidths.
We apply per-vector scaled QAT using a conventional QAT framework that leverages a straight-through estimator (STE) in the backward pass to propagate the gradient through any quantizer.
While the framework trains the weights that get fed into the quantizers in the model, the quantization scale factors are not parameters and are not explicitly trained.
Table~\ref{tbl:qat_study} evaluates the best accuracy achieved with QAT-based finetuning for both per-vector scaled quantization and per-channel scaled quantization. 
The number of retraining epochs taken to recover the specified accuracy is shown in parentheses.
Based on the presented cases in Table~\ref{tbl:qat_study}, per-vector scaled QAT gives significantly better accuracy than per-channel scaled QAT and requires much less effort to recover accuracy loss from quantization.

\begin{table}[t]
\vspace{5pt}
\begin{center}
\small
 \begin{tabular}{c|c|c c} 
 \hline
 \multirow{2}{*}{Model} & \multirow{2}{*}{Bitwidths} & \multicolumn{2}{c}{Accuracy with QAT} \\ \cline{3-4}
 & & PVAW & POC \\ \hline\hline
 ResNet50 & Wt=3 Act=3U & 75.53 (20) & 72.02 (20) \\ \hline 
 \multirow{2}{*}{BERT-base} & Wt=4 Act=4 & 86.93 (10) & 41.45 (20) \\ 
 & Wt=4 Act=8 & 87.80 (2) & 87.01 (2) \\ \hline  
 \multirow{2}{*}{BERT-large} & Wt=3 Act=4 & 89.26 (2) & 21.61 (10) \\
 & Wt=3 Act=8 & 90.59 (1) & 88.34 (1) \\ \hline
\end{tabular}
\vspace{-3pt}
\caption{\BF{QAT study --} Compares the best accuracy achieved after QAT-based finetuning. The number of retraining epochs taken to recover the accuracy is shown in parentheses.}
\label{tbl:qat_study}
\vspace{3pt}
\end{center}
\end{table}
\section{Conclusions}
\label{sec-conclusions}

In this paper, we introduced {\em VS-Quant}, a novel per-vector scaled quantization technique that employs per-vector scale factors to mitigate accuracy loss typical in existing quantized DNN models.
To support efficient per-vector scaling in hardware, we implemented a two-level scaling scheme and associated algorithm that combine a set of fine-grained scale factors with each coarse-grained scale factor.
We evaluated {\em VS-Quant} on a set of popular DNN models and tasks and demonstrated that it achieves significant improvement in post-training quantization accuracy when compared to conventional per-channel scaled quantization techniques. 

By extending the vector MAC unit of a DNN accelerator to dynamically support per-vector scaling at inference-time, we analyze the area and power implications of per-vector scaling on the hardware.
Experiments demonstrate that {\em VS-Quant} with 4-bit weights and activations achieves 37\% area saving and 24\% energy saving while maintaining over 75\% accuracy for ResNet50 on ImageNet.
Furthermore, {\em VS-Quant} with 4-bit weights and 8-bit activations achieves near-full-precision accuracy for both BERT-base and BERT-large on SQuAD while reducing area by 26\% compared to a non-{\em VS-Quant} 8-bit baseline.
By exploring the design space, we find that per-vector scaling provides better accuracy, energy, and area tradeoffs for low-precision inference.

For future work, we will continue to optimize the {\em VS-Quant} hardware and study the effect of scale factor and other intermediate rounding.
We will extend QAT to explicitly learn per-vector scale factors and co-optimize model architectures themselves with the {\em VS-Quant} hardware.

\clearpage

\bibliography{quantization}

\begin{thebibliography}{39}
\providecommand{\natexlab}[1]{#1}
\providecommand{\url}[1]{\texttt{#1}}
\expandafter\ifx\csname urlstyle\endcsname\relax
  \providecommand{\doi}[1]{doi: #1}\else
  \providecommand{\doi}{doi: \begingroup \urlstyle{rm}\Url}\fi

\bibitem[Bhandare et~al.(2019)Bhandare, Sripathi, Karkada, Menon, Choi, Datta,
  and Saletore]{bhandare2019efficient}
Bhandare, A., Sripathi, V., Karkada, D., Menon, V., Choi, S., Datta, K., and
  Saletore, V.
\newblock {Efficient 8-bit Quantization of Transformer Neural Machine Language
  Translation Model}.
\newblock \emph{arXiv preprint arXiv:1906.00532}, 2019.

\bibitem[Cai et~al.(2020)Cai, Yao, Dong, Gholami, Mahoney, and
  Keutzer]{cai2020zeroq}
Cai, Y., Yao, Z., Dong, Z., Gholami, A., Mahoney, M.~W., and Keutzer, K.
\newblock {ZeroQ: A Novel Zero Shot Quantization Framework}.
\newblock \emph{Conf. on Computer Vision and Pattern Recognition (CVPR)}, 2020.

\bibitem[Choi et~al.(2018)Choi, Wang, Venkataramani, Chuang, Srinivasan, and
  Gopalakrishnan]{choi2018pact}
Choi, J., Wang, Z., Venkataramani, S., Chuang, P. I.-J., Srinivasan, V., and
  Gopalakrishnan, K.
\newblock {PACT: Parameterized Clipping Activation for Quantized Neural
  Networks}.
\newblock \emph{arXiv preprint arXiv:1805.06085}, 2018.

\bibitem[Courbariaux et~al.(2014)Courbariaux, Bengio, and
  David]{courbariaux2014training}
Courbariaux, M., Bengio, Y., and David, J.-P.
\newblock {Training Deep Neural Networks with Low Precision Multiplications}.
\newblock \emph{arXiv preprint arXiv:1412.7024}, 2014.

\bibitem[Courbariaux et~al.(2015)Courbariaux, Bengio, and
  David]{courbariaux2015binaryconnect}
Courbariaux, M., Bengio, Y., and David, J.-P.
\newblock {BinaryConnect: Training Deep Neural Networks with Binary Weights
  during Propagations}.
\newblock \emph{Conf. on Neural Information Processing Systems (NeurIPS)},
  2015.

\bibitem[Fang et~al.(2020)Fang, Shafiee, Abdel-Aziz, Thorsley, Georgiadis, and
  Hassoun]{fang2020near}
Fang, J., Shafiee, A., Abdel-Aziz, H., Thorsley, D., Georgiadis, G., and
  Hassoun, J.
\newblock {Near-lossless Post-training Quantization of Deep Neural Networks via
  a Piecewise Linear Approximation}.
\newblock \emph{arXiv preprint arXiv:2002.00104}, 2020.

\bibitem[Gong et~al.(2014)Gong, Liu, Yang, and Bourdev]{gong2014compressing}
Gong, Y., Liu, L., Yang, M., and Bourdev, L.
\newblock {Compressing Deep Convolutional Networks using Vector Quantization}.
\newblock \emph{arXiv preprint arXiv:1412.6115}, 2014.

\bibitem[Gray(1984)]{gray1984vector}
Gray, R.
\newblock {Vector Quantization}.
\newblock \emph{IEEE ASSP Magazine}, 1984.

\bibitem[Han et~al.(2015)Han, Mao, and Dally]{han2015deep}
Han, S., Mao, H., and Dally, W.~J.
\newblock {Deep Compression: Compressing Deep Neural Networks with Pruning,
  Trained Quantization and Huffman Coding}.
\newblock \emph{arXiv preprint arXiv:1510.00149}, 2015.

\bibitem[Hubara et~al.(2016)Hubara, Courbariaux, Soudry, El-Yaniv, and
  Bengio]{hubara2016binarized}
Hubara, I., Courbariaux, M., Soudry, D., El-Yaniv, R., and Bengio, Y.
\newblock {Binarized Neural Networks}.
\newblock \emph{Conf. on Neural Information Processing Systems (NeurIPS)},
  2016.

\bibitem[Jain et~al.(2019)Jain, Venkataramani, Srinivasan, Choi,
  Gopalakrishnan, and Chang]{biscaled_dnn}
Jain, S., Venkataramani, S., Srinivasan, V., Choi, J., Gopalakrishnan, K., and
  Chang, L.
\newblock {BiScaled-DNN: Quantizing Long-tailed Datastructures with Two Scale
  Factors for Deep Neural Networks}.
\newblock \emph{Design Automation Conf. (DAC)}, 2019.

\bibitem[Khoram \& Li(2018)Khoram and Li]{khoram2018adaptive}
Khoram, S. and Li, J.
\newblock {Adaptive Quantization of Neural Networks}.
\newblock \emph{Int'l Conf. on Learning Representations (ICLR)}, 2018.

\bibitem[K{\"o}ster et~al.(2017)K{\"o}ster, Webb, Wang, Nassar, Bansal,
  Constable, Elibol, Gray, Hall, Hornof, et~al.]{koster2017flexpoint}
K{\"o}ster, U., Webb, T., Wang, X., Nassar, M., Bansal, A.~K., Constable, W.,
  Elibol, O., Gray, S., Hall, S., Hornof, L., et~al.
\newblock {Flexpoint: An adaptive Numerical Format for Efficient Training of
  Deep Neural Networks}.
\newblock \emph{Conf. on Neural Information Processing Systems (NeurIPS)},
  2017.

\bibitem[Krishnamoorthi(2018)]{krishnamoorthi2018quantizing}
Krishnamoorthi, R.
\newblock {Quantizing Deep Convolutional Networks for Efficient Inference: A
  Whitepaper}.
\newblock \emph{arXiv preprint arXiv:1806.08342}, 2018.

\bibitem[LeCun et~al.(2015)LeCun, Bengio, and Hinton]{lecun2015deep}
LeCun, Y., Bengio, Y., and Hinton, G.
\newblock {Deep Learning}.
\newblock \emph{Nature}, 521\penalty0 (7553):\penalty0 436--444, 2015.

\bibitem[Lee et~al.(2018)Lee, Ha, Choi, Lee, and Lee]{lee2018quantization}
Lee, J.~H., Ha, S., Choi, S., Lee, W.-J., and Lee, S.
\newblock {Quantization for Rapid Deployment of Deep Neural Networks}.
\newblock \emph{arXiv preprint arXiv:1810.05488}, 2018.

\bibitem[McKinstry et~al.(2018)McKinstry, Esser, Appuswamy, Bablani, Arthur,
  Yildiz, and Modha]{mckinstry2018discovering}
McKinstry, J.~L., Esser, S.~K., Appuswamy, R., Bablani, D., Arthur, J.~V.,
  Yildiz, I.~B., and Modha, D.~S.
\newblock {Discovering Low-precision Networks Close to Full-precision Networks
  for Efficient Embedded Inference}.
\newblock \emph{arXiv preprint arXiv:1809.04191}, 2018.

\bibitem[Mishra \& Marr(2017)Mishra and Marr]{mishra2017apprentice}
Mishra, A. and Marr, D.
\newblock {Apprentice: Using Knowledge Distillation Techniques to Improve
  Low-precision Network Accuracy}.
\newblock \emph{arXiv preprint arXiv:1711.05852}, 2017.

\bibitem[Miyashita et~al.(2016)Miyashita, Lee, and
  Murmann]{miyashita2016convolutional}
Miyashita, D., Lee, E.~H., and Murmann, B.
\newblock {Convolutional Neural Networks using Logarithmic Data
  Representation}.
\newblock \emph{arXiv preprint arXiv:1603.01025}, 2016.

\bibitem[Moons et~al.(2017)Moons, Goetschalckx, Van~Berckelaer, and
  Verhelst]{moons2017minimum}
Moons, B., Goetschalckx, K., Van~Berckelaer, N., and Verhelst, M.
\newblock {Minimum Energy Quantized Neural Networks}.
\newblock \emph{Asilomar Conference on Signals, Systems, and Computers}, 2017.

\bibitem[Nagel et~al.(2019)Nagel, Baalen, Blankevoort, and
  Welling]{nagel2019data}
Nagel, M., Baalen, M.~v., Blankevoort, T., and Welling, M.
\newblock {Data-free Quantization through Weight Equalization and Bias
  Correction}.
\newblock \emph{Int'l Conf. on Computer Vision (ICCV)}, 2019.

\bibitem[{NVIDIA Corporation}(2020)]{ampere3}
{NVIDIA Corporation}.
\newblock {NVIDIA A100 Tensor Core GPU Architecture}.
\newblock \emph{NVIDIA}, 2020.

\bibitem[Prato et~al.(2019)Prato, Charlaix, and Rezagholizadeh]{prato2019fully}
Prato, G., Charlaix, E., and Rezagholizadeh, M.
\newblock {Fully Quantized Transformer for Improved Translation}.
\newblock \emph{arXiv preprint arXiv:1910.10485}, 2019.

\bibitem[Rouhani et~al.(2020)Rouhani, Lo, Zhao, Liu, Fowers, Ovtcharov,
  Vinogradsky, Massengill, Yang, Bittner, et~al.]{darvish2020pushing}
Rouhani, B., Lo, D., Zhao, R., Liu, M., Fowers, J., Ovtcharov, K., Vinogradsky,
  A., Massengill, S., Yang, L., Bittner, R., et~al.
\newblock {Pushing the Limits of Narrow Precision Inferencing at Cloud Scale
  with Microsoft Floating Point}.
\newblock \emph{Conf. on Neural Information Processing Systems (NeurIPS)},
  2020.

\bibitem[Shen et~al.(2020)Shen, Dong, Ye, Ma, Yao, Gholami, Mahoney, and
  Keutzer]{shen2020q}
Shen, S., Dong, Z., Ye, J., Ma, L., Yao, Z., Gholami, A., Mahoney, M.~W., and
  Keutzer, K.
\newblock {Q-BERT: Hessian Based Ultra Low Precision Quantization of BERT}.
\newblock \emph{AAAI Conf. on Artificial Intelligence (AAAI)}, 2020.

\bibitem[Sijstermans(2018)]{sijstermans2018nvdla}
Sijstermans, F.
\newblock {The NVIDIA Deep Learning Accelerator}.
\newblock \emph{Symp. on High Performance Chips (Hot Chips)}, 2018.

\bibitem[Stock et~al.(2020)Stock, Joulin, Gribonval, Graham, and
  J{\'e}gou]{stock2020and}
Stock, P., Joulin, A., Gribonval, R., Graham, B., and J{\'e}gou, H.
\newblock {And the Bit Goes Down: Revisiting the Quantization of Neural
  Networks}.
\newblock \emph{Int'l Conf. on Learning Representations (ICLR)}, 2020.

\bibitem[Sze et~al.(2020)Sze, Chen, Yang, and Emer]{sze2020efficient}
Sze, V., Chen, Y.-H., Yang, T.-J., and Emer, J.~S.
\newblock {Efficient Processing of Deep Neural Networks}.
\newblock \emph{Synthesis Lectures on Computer Architecture}, 2020.

\bibitem[Tambe et~al.(2020)Tambe, Yang, Wan, Deng, Reddi, Rush, Brooks, and
  Wei]{tambealgorithm}
Tambe, T., Yang, E.-Y., Wan, Z., Deng, Y., Reddi, V.~J., Rush, A., Brooks, D.,
  and Wei, G.-Y.
\newblock {Algorithm-Hardware Co-Design of Adaptive Floating-Point Encodings
  for Resilient Deep Learning Inference}.
\newblock \emph{Design Automation Conference (DAC)}, 2020.

\bibitem[Venkatesan et~al.(2019)Venkatesan, Shao, Wang, Clemons, Dai, Fojtik,
  Keller, Klinefelter, Pinckney, Raina, et~al.]{venkatesan2019magnet}
Venkatesan, R., Shao, Y.~S., Wang, M., Clemons, J., Dai, S., Fojtik, M.,
  Keller, B., Klinefelter, A., Pinckney, N.~R., Raina, P., et~al.
\newblock {MAGNet: A Modular Accelerator Generator for Neural Networks.}
\newblock \emph{Int'l Conf. on Computer Aided Design (ICCAD)}, 2019.

\bibitem[Wu et~al.(2018)Wu, Wang, Zhang, Tian, Vajda, and Keutzer]{wu2018mixed}
Wu, B., Wang, Y., Zhang, P., Tian, Y., Vajda, P., and Keutzer, K.
\newblock {Mixed Precision Quantization of ConvNets via Differentiable Neural
  Architecture Search}.
\newblock \emph{arXiv preprint arXiv:1812.00090}, 2018.

\bibitem[Wu(2019)]{wu2019low}
Wu, H.
\newblock {Low Precision Inference on GPUs}.
\newblock \emph{GPU Technology Conference (GTC)}, 2019.

\bibitem[Wu et~al.(2020)Wu, Judd, Zhang, Isaev, and
  Micikevicius]{wu2020integer}
Wu, H., Judd, P., Zhang, X., Isaev, M., and Micikevicius, P.
\newblock {Integer Quantization for Deep Learning Inference: Principles and
  Empirical Evaluation}.
\newblock \emph{arXiv preprint arXiv:2004.09602}, 2020.

\bibitem[Wu et~al.(2016{\natexlab{a}})Wu, Leng, Wang, Hu, and
  Cheng]{wu2016quantized}
Wu, J., Leng, C., Wang, Y., Hu, Q., and Cheng, J.
\newblock {Quantized Convolutional Neural Networks for Mobile Devices}.
\newblock \emph{Conf. on Computer Vision and Pattern Recognition (CVPR)},
  2016{\natexlab{a}}.

\bibitem[Wu et~al.(2016{\natexlab{b}})Wu, Schuster, Chen, Le, Norouzi,
  Macherey, Krikun, Cao, Gao, Macherey, et~al.]{wu2016google}
Wu, Y., Schuster, M., Chen, Z., Le, Q.~V., Norouzi, M., Macherey, W., Krikun,
  M., Cao, Y., Gao, Q., Macherey, K., et~al.
\newblock {Google's Neural Machine Translation System: Bridging the Gap between
  Human and Machine Translation}.
\newblock \emph{arXiv preprint arXiv:1609.08144}, 2016{\natexlab{b}}.

\bibitem[Zafrir et~al.(2019)Zafrir, Boudoukh, Izsak, and
  Wasserblat]{zafrir2019q8bert}
Zafrir, O., Boudoukh, G., Izsak, P., and Wasserblat, M.
\newblock {Q8bert: Quantized 8bit BERT}.
\newblock \emph{arXiv preprint arXiv:1910.06188}, 2019.

\bibitem[Zhao et~al.(2019)Zhao, Hu, Dotzel, De~Sa, and
  Zhang]{zhao2019improving}
Zhao, R., Hu, Y., Dotzel, J., De~Sa, C., and Zhang, Z.
\newblock {Improving Neural Network Quantization without Retraining using
  Outlier Channel Splitting}.
\newblock \emph{Int'l Conf. on Machine Learning (ICML)}, 2019.

\bibitem[Zhou et~al.(2016)Zhou, Wu, Ni, Zhou, Wen, and Zou]{zhou2016dorefa}
Zhou, S., Wu, Y., Ni, Z., Zhou, X., Wen, H., and Zou, Y.
\newblock {DoReFa-net: Training Low Bitwidth Convolutional Neural Networks with
  Low Bitwidth Gradients}.
\newblock \emph{arXiv preprint arXiv:1606.06160}, 2016.

\bibitem[Zhu et~al.(2016)Zhu, Han, Mao, and Dally]{zhu2016trained}
Zhu, C., Han, S., Mao, H., and Dally, W.~J.
\newblock {Trained Ternary Quantization}.
\newblock \emph{arXiv preprint arXiv:1612.01064}, 2016.

\end{thebibliography}
\bibliographystyle{mlsys2020}

\end{document}